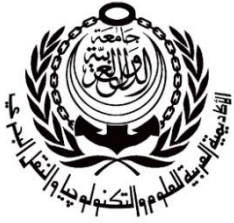

**ARAB ACADEMY FOR SCIENCE, TECHNOLOGY, AND MARITIME TRANSPORT**

**College of Computing & Information Technology –Cairo Campus**

**Department of Computer Science**

# A MULTILINGUAL ENCODING METHOD FOR TEXT CLASSIFICATION AND DIALECT IDENTIFICATION USING CONVOLUTIONAL NEURAL NETWORK

**By**

*Amr Adel Helmy Amr*

**Egypt**

**A thesis submitted to the College of Computing & Information Technology in partial fulfillment of the requirements for the award of the degree of**

**MASTER OF SCIENCE**

**In**

**Computer Science**

## Supervisors

| | |
|---|---|
| **Ass. Prof. Yasser M.K. Omar** | **Ass. Prof. Rania Hodhod** |
| Computing and Information Technology | TSYS School of Computer Science, |
| Arab Academy for Science, | Columbus State University, |
| Technology, and Maritime Transport | Columbus, GA |
| Cairo, Egypt | USA |
**February 2019**
**Amr Adel Helmy**
Email: aaheg@msn.com

# Abstract

Text classification plays a vital role today especially with the intensive use of social networking media. Recently, convolutional neural networks have been used for text classification in which one-hot vector and word-embedding methods are commonly used. This thesis presents a new language-independent word encoding method for text classification, where a word is an atomic representation of the document as in word level. In this method, raw text data is converted to a low-level feature dimension, with minimal preprocessing steps by using two new approaches called binary unique number of word "BUNOW" and binary unique number of character "BUNOC." BUNOW and BUNOC allow each unique N-gram to have an integer ID in a dictionary that can be represented as a k-dimensional vector of its binary equivalent. The output vector of this encoding is fed into a convolutional neural network (CNN) for classification. The proposed method is able to directly reduce the neural network parameters and the memory consumption as in character level representation, in addition to providing much faster computations with few network layers. Our models have been evaluated in two different morphological languages by using two-benchmarked dataset one for Arabic language and one for English language. The provided CNN model outperforms the character level and very deep character level CNNs in terms of accuracy, network parameters, and memory consumption. In English AG's dataset, the proposed model allows for reduction in input feature vector and neural network parameters by 62% and 34%, respectively. Results from testing the model show total classification accuracy 91.99%, and error 8.01% compared to the state of art methods that have total classification accuracy 91.45% and error 8.55%. In Arabic AOC dataset, the proposed model achieved competitive accuracy results of 81.59% in validation and 79.76% in testing compared to the best result of 80.15% and 78.61 achieved from 7 traditional machine-learning classifiers, and 12 different deep learning architectures. Moreover, the model proved to be general enough to work with other languages or multi-lingual text without the need for any changes in the encoding method.

# TABLE OF CONTENTS

# List of Tables

# List of Figures

# List of Abbreviations

| | |
|---|---|
| AG's | English News Dataset |
| ANN | Artificial Neural Network |
| AOC | Arabic Online Commentary Dataset |
| Bi | Bidirectional |
| BOW | Bag of Words |
| CBOW | Continuous Bag of Words |
| CLSTM | Convolutional Long Short Time Memory |
| CNN | Convolutional Neural Network |
| DL | Deep Learning |
| DNN | Deep Neural Network |
| GPU | Graphical Processing Unit |
| GRU | Gated Recurrent Unit |
| IG | Information Gain |
| LSTM | Long Short Time Memory |
| NLP | Natural Language Processing |
| NN | Neural Network |
| OOV | Out-Of-Vocabulary |
| RELU | Rectified Linear Unit |
| ResNets | Residual Neural Network |
| SVM | Support Vector Machine |
| UTF-8 | Unicode Transformation Format - 8 Bit |
| VGG | Visual Geometry Group Neural Model |
| WE | Word Embedding |

# Chapter 1: Introduction

Text classification is one of the essential tasks in Natural Language Processing (NLP). It is the task of automatically assigning pre-defined categories to documents written in natural languages. Text classification can be used for classifying a document according to its topics, rating a product review by customer or analysis a text sentiment opinion of a user toward a product or a service [1] [2] [3].

Traditional machine learning techniques commonly used to build a text classification model in which a domain specialist struggles to choose the best feature selection criteria for each specific classification task. The specialist usually relies on handcrafted features selection such as n-grams, negation words, stop words, punctuation, emoticons, elongated words, and lexicons in order to achieve a good classification performance. This approach is called feature engineering [4].

## 1.1 Research Motivation

We are living in the data age in which internet evolution has become the most popular communication platforms for public reviews, opinions, comments, recommendations, ratings, feedback, attitudes, emotions, and feelings. This can be about products, places, books, research papers, application, websites, and social.

Powerful and versatile tools are badly needed to automatically discover valuable information from this explosively growing, widely available, and gigantic body of data, and to transform such data into organized knowledge. This huge amount of valuable data directly affects the businesses, as it becomes one of the most important sources for decision making, which encourages the researcher and decision makers to analyze it.

The amount of data increases every minute especially text (text sent-email spam-Google search …etc.) as shown in Figure 1.

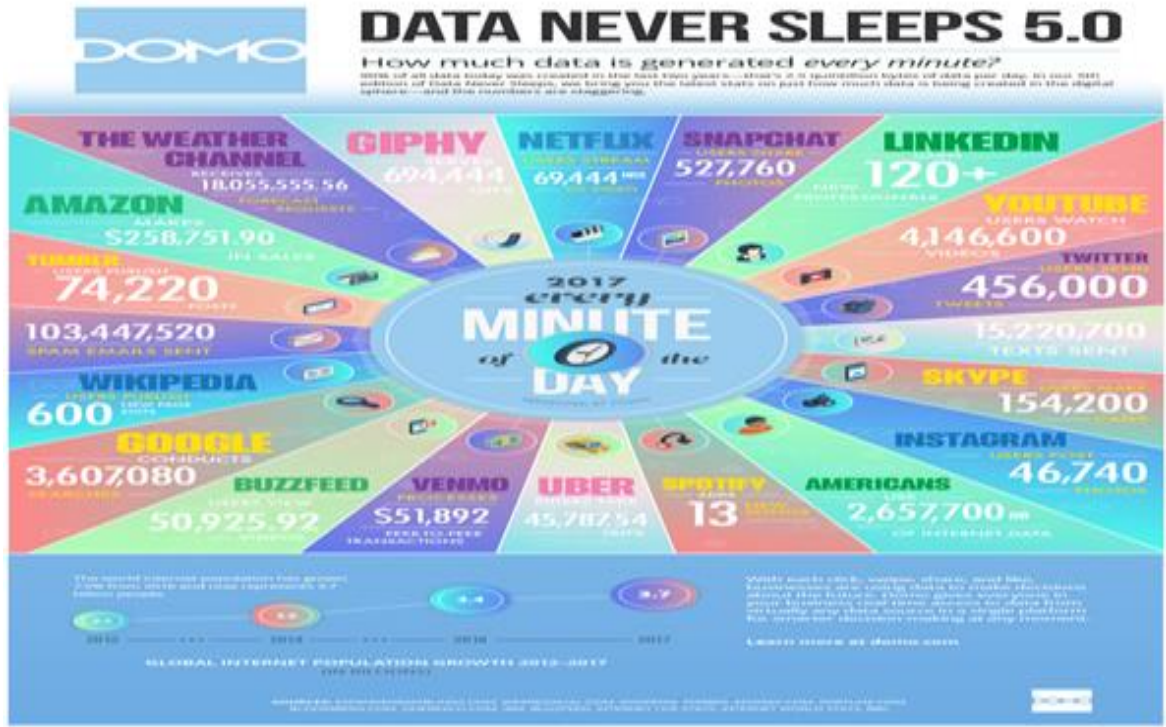

<p align="center">**Figure 1: Data Never Sleeps 5.0** [5]</p>

## 1.2 Problem Statement

Text classification uses natural language processing (NLP) to classify text data to its predefined categories. Machine Learning (ML) techniques are intensively used for this task because of their ability to automatically recognize the different complex patterns and distinguish between them.

Text data is characterized by very high dimensionality that can cause a phenomenon called curse of dimensionality [6]. This makes it unsuitable for training using machine learning techniques without applying feature selection methodologies in order to extract meaningful features that can help reducing the dimensionality. Accordingly, feature extraction and efficient representation of text become important factors in improving the accuracy of classification. Many text representation methods have been used, such as bag-of-words (BOWs) [7], Term Frequency, and Inverse Document Frequency (TF-IDF) [8], and Latent Semantic Indexing (LSI) [9] in order to select the appropriate features to be used by classifiers like Naïve Bayes (NB), k-nearest neighbors (KNN), and support vector machine (SVM).

One of the critical questions posed by researchers in automatic text classification is how to choose the best feature vector for traditional machine learning classifiers? The challenge lies in the fact that there is no single feature engineering technique that can work for all text classification and learning tasks. In other words, there is no best practice known in this area. In general, feature engineering has its drawbacks due to loss of information, needs manual fine-tuning of data, and requires prior familiarity with the language. An alternative way is to use deep learning for automatically learn features from raw text data.

Recently, CNNs have proved to be well suited for text classification as it outperforms other models, such as bag-of-words (BOWs). CNNs proved its ability to learn automatically from scratch using character-level representations of text irrespectively of language used and without prior knowledge of language words, syntax, grammar, and semantic similarities [10]. CNNs allow high-level understanding provided the availability of sufficient data.

The main drawback of CNNs in text classification is the extensive use of time and memory, in addition to expensive computation and resources consumption especially with large volume of training data with large word vocabulary size.

## 1.3 Research Objectives

The use of word embedding (Word2Vec) technique with CNNs has attracted considerable attention from many researchers over the traditional models, such as One-hot vector and BOW because of its ability to reduce memory requirements and training time, in addition to its effect on performance. In the case of BOW, each word is represented as one-hot vector with dimensions equal to the words vocabulary size (N); each 1 digit is placed in the correspondent position of that word in a 1-N vector, all other positions are filled with 0 digits. Since natural languages are characterized by huge vocabulary size, it is common to use word frequency statistics or relevance metrics to determine the most frequent words that are representative of the texts and exclude the rare ones. This helps to control the vocabulary size that directly affects the computational and memory requirements.

In the Word2Vec approach, each word is projected into an embedding metric of fixed size that represents its co-occurrence in a text corpus; The authors in [11] used two widely architectures model, CBOW model and the skip-gram model to compute continuous vector representations of words from very large data sets. Both models use dense vectorization to represent the word vectors. Their work provided cutting-edge results for measuring syntactic and semantic word similarities.

This thesis proposes a new language-independent text representation technique for feeding text data into a single hybrid character-word CNN classification model. The model achieved high accuracy results compared to the state of the art of CNN models. The proposed model takes advantage of character level in terms of less preprocessing for raw data, less features vector dimension, less neural network parameters, in addition to word level in terms of using fewer convolutional layers, where a word is an atomic representation of the document.

## 1.4 Organization of the Thesis

This thesis is organized as follows. Background overview is presented in the next Chapter. A literature review discussion is given in Chapter 3. The proposed model architecture is described in detail in Chapter 4. Results compared to the state of the art in text classification and dialect identification with experiment methodology are detailed in Chapter 5. Discussion of thesis results is summarized in Chapter 6. Finally, conclusions and future work are presented in Chapter 7.

# Chapter 2: Background

In this chapter, we present an overview and the state of the art of the techniques used in this work.

## 2.1 Natural Language Processing

Natural language processing (NLP) also known as computational linguistics is a multidiscipline field that has been influenced by other fields such as linguistics, psychology, philosophy, cognitive science, probability and statistics, and machine learning. NLP is the process of understanding human languages through addressing the fundamental problems such as language modeling, morphological processing (dealing with segmentation of meaningful components of words, and identifying the true parts of speech of words as used), and syntactic processing or parsing that attempts to distill meaning of words, phrases, and higher level components in a piece of text.

NLP is involved in many application areas such as extraction of useful information from text (e.g., named entities and relations), translation of text, and summarization of written works, automatic answering of questions by inferring the most probable answers, and classification, and clustering of documents.

Early attempts at NLP were usually rule based, where rules were handcrafted using knowledge derived from various areas. Sometimes the rules were ad-hoc in that they were made up to solve specific problems expediently. A number of formalisms were developed to describe the syntax of natural languages, with a view toward the facilitation of parsing and semantic processing. Syntactic rules were usually paired with logic-based semantic rules to obtain semantic representations of sentences that were then used for solving practical problems. Early approaches to NLP were described in depth in widely used textbooks of the time [12] [13].

Since1980s, rule base approaches have become insufficient to handle the informal language used at social media, texting on phones, and the democratization of writing on the Web, this is attributed to humans do not follow any language rules of spelling and grammar when they write or speak. NLP began to transform slowly towards data-driven approaches [14] [15] [1] by using statistical and probabilistic computations along with machine learning algorithms in order to face the new challenge of informal language. For example, parsing, which used to be rule based, has become driven by statistical computation and machine learning, especially after being supported by corpora containing large numbers of parse trees, such as the Penn Treebank [16]. Additionally, large corpora of carefully collected texts of various kinds have become available, ranging from large multi-lingual corpora of well-written formal and legal documents, to collections of pages obtained from Wikipedia, all the way to sets of highly informal texts such as tweets and other social media posts.

Over time, a number of machine learning approaches such as Naïve Bayes, k-nearest neighbors, hidden Markov models, conditional random fields, decision trees, random forests, and support vector machines were widely used in NLP. However, during the past several years, many of these approaches have been entirely replaced, or at least enhanced, by neural models such as the works done in [4] [10] [17] [18].

## 2.2 Text Classification

Text classification is one of main task in natural language processing field [19]. It is the process of associating a text to its predefined category or label depends on the content of the text.

The rapidly increasing in text volume due to intensively using of internet for text communication especially in social media, expresses opinion or comments toward product or services make it difficult task to handle with existence methods. This encourages the researchers to looking after a new statistical and artificial intelligence tools for automating this process. An important step in this process is how to representing the text.

The early works relying on a concept of Bag of Words [20] where the numbers of times a word exist in training dataset has a big influence in selecting this word. Although this model is able to classify the text based on word count but it is unable to distinguish between two class article like sport and news especially when one of those class have a much more frequent word appearance than other. That is why the inverse document frequency term added to word count to perform a TF-IDF method to overcome this drawback.

The N-gram concept is commonly used in text classification for document representation and can be seen as an extension version of the Bag of Word model [21]. The N-gram model selects the feature based on the most frequent N-continuous word combinations from the training dataset. It is widely used in natural language processing due to its simplicity and powerful in small dataset.

### 2.2.1 Artificial Neural Network

The artificial neural network (ANN) is a special type of machine learning algorithms and one of its popular algorithms inspired by the biological neural networks, which was discovered by neuroscience. ANN started in the 1950s in order to develop computer programs which capable of solving abstract problems as easily as human does [22] [23].

The human neural network, shown in Figure 2, consists of cortex, which is estimated to have 10 billion neurons and 60 trillion synapses [24] and can be viewed as a complex network whose nodes are neurons. Each neuron sends its output signal to other neurons through its dendrites and connections are established through the synapses. The strength of the signals is determined by the stimulus received. The output of a neuron signal is called a spike and is created by combining the entire input signal received from other neurons and transmitted to other neurons through the axon if its amplitude is greater than a pre-determined value called action potential.

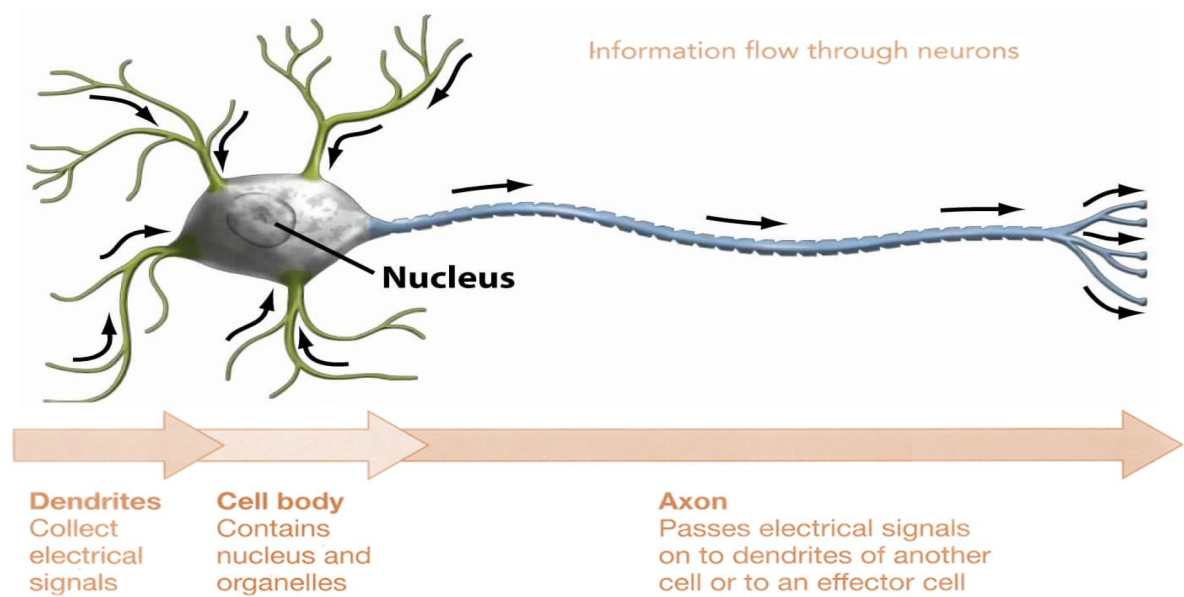

**Figure 2: Neural Network Architecture**

The scholars in [25] built an artificial neural network architecture for storing experimental knowledge which consist of a massively parallel distributed system with simple processing units, similar to human neural network.

The learning mechanism in an ANN depends mainly on synaptic weights, or network parameters adjustments through a stimulation process known as training. An error-correction algorithm is commonly used to compare the network output with a target value by using a cost function associated with network parameters to measure the error produced by the network output, and determine the way the parameters changes take place.

Back-propagation, one of the most used error-correction learning algorithm in feed-forward ANNs, is used to adjust parameter weight which was originally introduced in the 1970s, and it became popular only in 1986 after the publication of a paper [26] that proved that this algorithm enable ANNs to solve problems that had previously been insolvable. Its core idea has four steps process: 1. **In the feed-forward** step, the network output is calculated. 2. **In the error step**, the cost function gradient of with respect to the network output is calculated. 3. **In the backward step**, the error is back propagated calculating the gradient with respect to the previous layers' outputs. 4. **In the update step**, the values of the network parameters are adjusted using some updating rule (i.e. gradient descent algorithm) in which gradient multiplied by network parameters are subtracted by a constant predefined value called the learning rate.

### 2.2.2 Deep Learning

Deep learning (DL) is a branch of machine learning methods based on learning data. Deep learning architectures that involve Artificial Neural Networks (ANNs) are referred to as deep neural networks (DNNs) [4] [27]. They consist of many layers of information processing units, arranged in a hierarchical architecture to perform pattern classification and automatic feature learning [28].

A DNN is a powerful framework with complex structure not just a simple neural network that uses multiple hidden layers (i.e. Recurrent Neural Network (RNN) [29] and Convolutional Neural Network (CNN) [30]). In 2006, with the enhancement of the algorithms and the popularity of powerful GPUs for scientific computation task, the DNN models started to appear [31] [32]. Since then, DNNs have been re-emerged as an exciting research area, attracting a wide variety of scientists and engineers and have been successfully applied to a variety of NLP tasks.

DL had achieved an outstanding performance in various pattern-recognition, and machine-learning tasks in addition to its contribution in enhancing the work in advanced artificial intelligence projects, such as the self-driven car [33] compared to traditional machine learning algorithms. Furthermore, DL has been increasingly used for NLP and not just for machine learning tasks [10] [18] opening a new emerging research area in both of ML and NLP.

Over the previous several years, most natural language processing tasks (e.g., language identification, language modeling, chunking, part of speech tagging, morphological analysis) extensively rely on a number of models, such as SVM [34], CRF [35], Logistic Regression, Hidden Markov Models (HMM), and Perceptron [36]. With the advent of social media and big data, these traditional models became insufficient to process these huge amounts of textual information and handle the new challenge in oppose to DL techniques that have been shown a great promise future in solving many NLP challenges.

DNNs theoretically exist since long time and are known as universal non-linear function approximators [37]. During the period from 1990 until 2000, deep architectures were not widely spread because of their weakly performance; they were difficult to train in an efficient manner compared to the widely used shallow architectures that rely on feature engineering. It actually starts to exist after the work done by [38]. As they prevent DNN from staying stuck in a local minimum during back-propagation by introducing Deep Belief Networks (DBNs). DBNs are composed of simple learning units, particularly Restricted Boltzmann Machines (RBMs), in which DBNs weights are initialized by using unsupervised pre-trained learning algorithm in a greedy layer wise fashion.

Recently, the enhancement of computational power and parallelization harnessed by Graphical Processing Units (GPUs) [39] [40] allow the application of deep learning concept by utilizes ANNs with billions of trainable parameters [41]. Additionally, the availability of large datasets enables the training of such deep architectures via their associated learning algorithms [33] [42] [43].

### 2.2.3 Convolutional Neural Network

ANN has much architecture in which each one has been developed for a specific task. The Convolutional Neural Network (CNN) architecture idea introduced for image processing task based on neurophysiologists discovered about how the mammalian vision system works [44] which was studied the reaction of neurons inside cat's brain when a projected image shown to cat.

The receptive field of visual cortex consists of a complex arrangement of cells that are sensitive to small sub-regions of the visual field. The sub-regions are tiled to cover the whole visual field. The cells work as a local filter over the input space and easily capture the spatially correlation present in natural images.

Convolutional mean the term convolution used in mathematical operation, which is a special type of linear operation. Convolutional networks are simply neural network that use convolution in place of general matrix multiplication in one of their layers [41] as shown below in Figure 3.

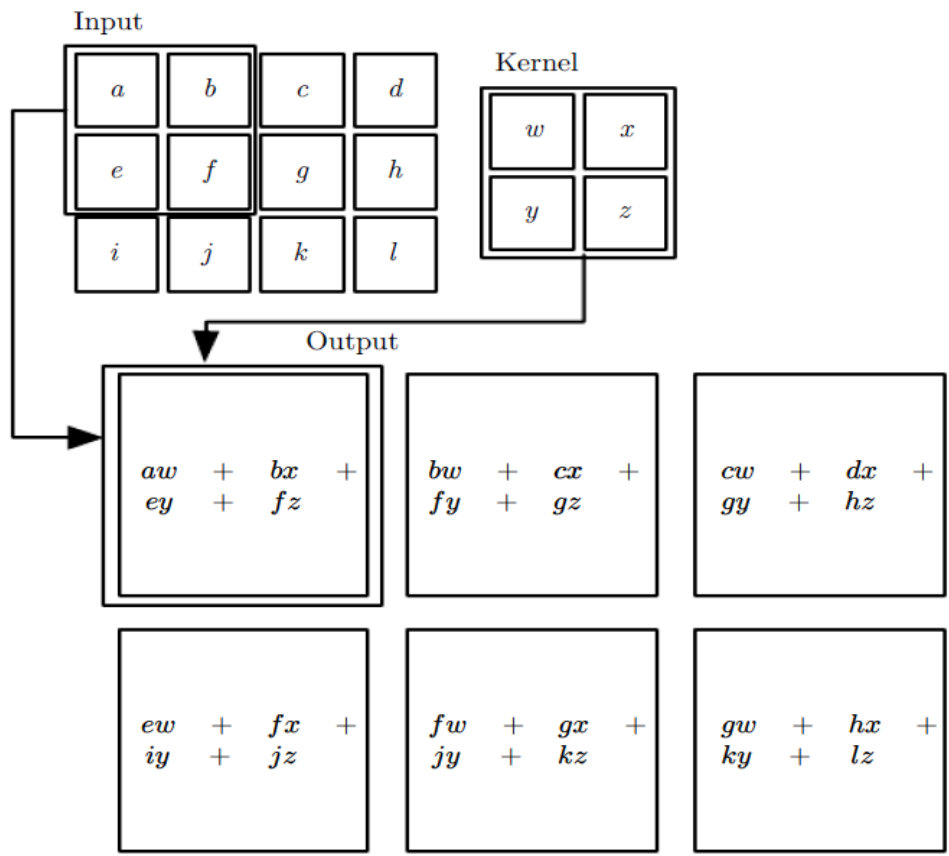

**Figure 3: Example of 2D Mathematical Convolution** [41]

The powerful of deep CNN models not only the ability of automatically extract the features from raw data, but also it is ability to learn a hierarchical representation of the data through the extracted features as shown in Figure 4. One of the key advantages of CNNs architecture over AANs is the way of connection between the layers, which required fewer parameters, and reducing memory consumption space in addition of its computing efficiency that required a fewer operation. These improvements in performance are usually quite large. For example, processing an image with thousands or millions of input pixels by traditional ANN layers, which use a matrix multiplication, required that every element of a layer be connected to every element of the previous and next layers. CNNs have sparse connections. This is accomplished by making the filter smaller than the input by detect small, meaningful features such as edges with filters that occupy only tens or hundreds of pixels. Another key advantage strategy present in CNNs is the parameter sharing that helps reduce the memory requirements. It is means the using of the same parameter for more than one function in a model. In oppose to a traditional ANN, in which each element of the weight matrix used exactly once when computing the output of a layer.

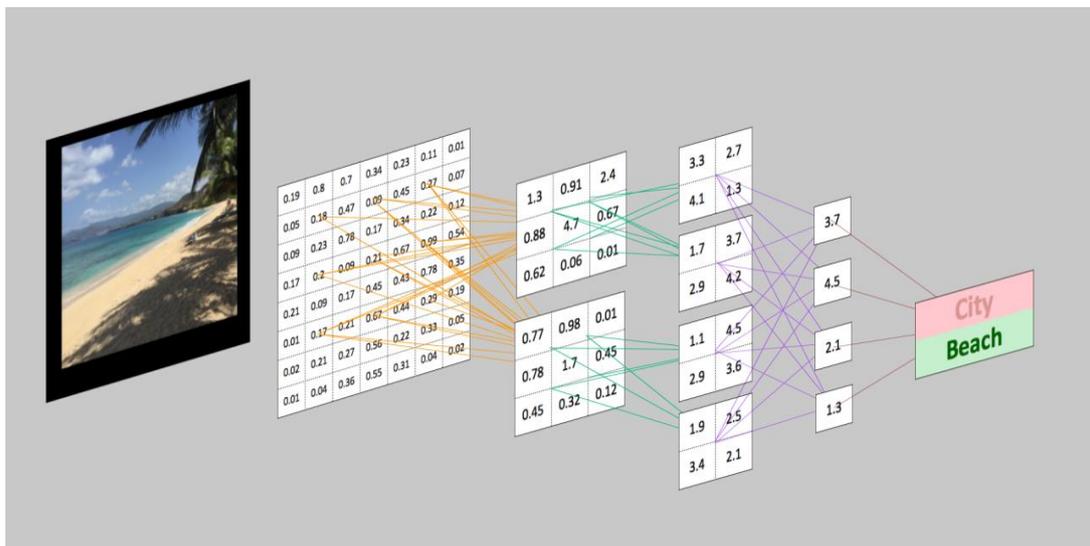

**Figure 4: An Example of CNN Image Recognition Task** [45]

A commonly used CNN model consists of three stages as shown in Figure 5. Firstly, the layer performs several parallel convolutions computation in order to produce a set of linear activations. Secondly, each linear activation is run through a nonlinear activation function, such as the rectified linear activation function. Thirdly, a pooling function used to modify the layer output at a certain location by the mean of statistic summarization of the nearby outputs.

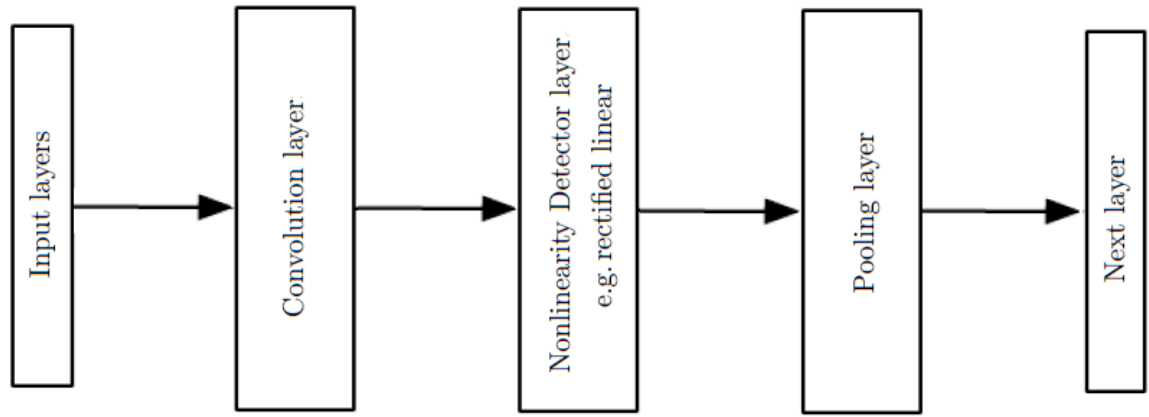

**Figure 5: Typical CNN Architecture**

Input layers are storing the raw input data for processing in the network. Convolutional Kernel Size is small spatial window with width and height smaller than the width and height of the input data. Convolutional Feature Maps is the output results of dot product between a sliding kernel size window with randomly generated weights and input data values. Stride control the horizontal and vertical sliding of kernel size window across input data.

The rectified linear unit (ReLU) layer is applying an element-wise activation function $g(z) = \max\{0, z\}$ over the input data, which change its value while preserve its spatial dimensions in output.

Pooling layers are commonly used for reduce the spatial size (width and height) of the data representation. The max-pooling operation detects the maximum output within a neighborhood and contributes in reducing the output of that layer before feeding to the next layer.

# Chapter 3: Literature Review

## 3.1 Related Work

### 3.1.1 English Text Classification

Text classification is one of the most important tasks in Natural Language Processing (NLP). Traditionally, the input features vector of a given document is represented as bag-of-words or n-grams where their frequency–inverse document frequency (TF-IDF) serves as the input for a subsequent linear classifier. The most popular classifiers used for text classification are support vector machines [46], Naive Bayes [47], and LIBLINEAR [48], and linear support vector machines [48].

Recently, deep learning methods, such as CNNs [10] [17] [18] are used for feature extraction and classification of text while bag-of-words (BOWs) [49] and Word2Vec [11] are used to represent textual information. The outputs from applying these methods serve as input to CNNs. Researchers in [17] proposed a shallow neural network with one convolutional layer (using multiple filter widths and feature maps), followed by a max over time pooling layer and one fully connected layer with dropout and softmax output. This convolutional layer works on top of initialized input word vectors from text data; each word is of k-dimensional size and is obtained from its correspondence pre-trained Word2Vec on 100 billion words of Google News [11]. Other researchers followed the same methodology of using CNN layers, in addition to introducing a max pooling layers with dynamic k size [50]. This assists in detecting the highly important k features in a text, regardless of their position, while taken in consideration their relative order (the network layer position and text length determine the value of k).

Many researchers have observed that it is not necessary to use the one-hot vector representation of word (word-level) with a deep neural network, character or even sub-word level can be used as an alternative. Researchers introduced the use of character sequence as a substitution to the word one-hot vector [51] and [52] because of its low representation vector as character depends only on the number of definite characters exist in the language used compared to huge vocabulary size associated with word representation [51] [52]. Character concept was also used instead of word for dependency parsing [53].

Other researchers noted that the word representation could be unsuitable for social media like Twitter, where tokens are usually a challenge due to the existence of slang, elongated words, and contiguous sequences of exclamation marks, and abbreviations, hash tags [54]. Accordingly, they introduced a characters-level CNN, which automatically learns the words and notions of sentences from scratch and without pre-processing or even tokenization. The most relevant work on character-level CNN for text classification was proposed by [10] (six CNN layers) and [18] (29 CNN layers).

More work was done on using CNN to train classifiers by representing text in an image like fashion where each character has an atomic representation [10]. This enabled a deep CNN to classify text with high-level concepts. In this work, Each English alphabetic character inside the document of length l (l is the length of sequences of character in the document) is encoded in one-hot vector (one-of-m) with value of one in the position of this character inside the vector, and 0 values for the rest, where m size is equivalent to alphabet vocabulary size. Using this encoding method, data is fed into a CNN model consists of six 1-D temporal CNN layers and three fully connected layers. Kernels size of three and seven are used in addition to simplest max-pooling layers. Features are extracted from small, overlapping windows of the input sequence of each layer and pools over small, non-overlapping windows by taking the maximum activations in the window. This method performs entirely at the character level and is able to learn from scratch with minimal processing and without prior knowledge of the language structure. This work shows that character-level CNN is an effective method.

The authors in [55] introduced a new character encoding method for binary tweet sentiment classification toward a dataset of 50,000 positive and 50,000 negative instances from the sentiment140 corpus. This idea is similar to Unicode UTF-8 of character representation by encoding each of the 70 character of English alphabet in a small model with a vector of size 7, and each of the 256 characters of the English alphabet in a large model with a vector of size 8. This encoding method reduces the input vector dimension from 1-of-70 in small models to 1-of-7 and from 1-of-256 in large models to 1-of-8. Three convolutional layers of 3x3-filter size, with a stride of one and no padding, followed by three fully connected layers, two dropout layers of 0.5 to avoid network over-fitting were used. This structure achieved a higher accuracy results compared to 1-of-m achieved of almost 3.3% using vocabulary size of 70 and 5.2% using vocabulary size of 256 also it allowed the training of the CNN to be 4.85 times faster.

The work presented in [56] introduced a CNN architecture with recurrent layers where a one-hot sequence input created from character vocabulary of ninety-six case sensitive characters, digit numbers, punctuation, and spaces is converted to a vector dense of size 8 using an embedding layer. The output of this layer is then fed into multiple convolutional layers of kernel size three or five depending on the depth and max-pooling layer of size two with rectified linear unit (ReLU) activation function to get a shorter feature vector [57]. This feature vector is then fed into a single recurrent layer with bidirectional Long-short term memory (LSTM) where it concatenates the last hidden state of both directions forming a fixed-dimensional vector. Finally, this vector is fed into the classification layer to compute the probabilities of each category. The scholars in this work applied dropout after the last convolutional layer to avoid deep neural networks over-fitting, and recurrent layer. This hybrid model was able to capture sub-word information from a character sequence in a document, and achieve comparable performances compared to [10].

A comparative study was presented in [58] to compare the accuracy of document sentiment classification on four different binary classes datasets with different characteristics toward support vector machine classifier with feature selection techniques, and deep learning model with word embedding techniques. For feature selection methods, information gain [59], Gini index [60], and distinguishing feature selector [61] were used to select the best BOW feature vector as an input for SVM classifier. While deep learning models, CNN, LSTM, and CNN+LSTM used pre-trained word embedding with and without tuning during training, and one-hot vector. In general, the deep learning models outperformed SVM with feature selection based approaches except when IG+WE used, it has outperformed all other models in only one dataset. The IG and DFS feature selection methods achieved higher accuracy than the GI method. Best accuracy result were achieved when deep learning models are used with either one-hot vectors or fine-tuned word embedding over word embedding without tuning method.

A new idea was presented in paper [62] depending on removing noisy data from a DBPedia dataset by filtering out uncommonly used words and incorrect spelled ones by the assistance of statistic methods, i.e. chi-square and term frequency-inverse document frequency (TF-IDF), or other dataset, such as Google News [11] in order to reduce large vocabulary size document which has a negative effect on accuracy. After cleaning the data, the vocabulary size of this dataset was limited to size 70K. The scholar implemented five-layer neural network in which the first layer is the embedding layer, with dimension 300, the length of sentence was fixed to 708, and batch size to 50. The second layer is a convolutional layer with three filters whose sizes are 3, 4, and 5. The number of feature maps is 100 for each size. The output shape of network was, embedded layer size of $708 \times 300$, and convolutional layer generated feature maps of size $706 \times 100$, $705 \times 100$ and $704 \times 100$ respectively. Then max-pooling layer is applied to the feature maps to generate an output dimensional shape of $1 \times 100$ per feature size, which are then concatenate together to produce a single matrix of $1 \times 300$. A fully connected layer with dropout of 0.5 was used such that the output connected to Softmax layer for classification label output.

Work done in [18] proposed a new character level CNN with very deep CNN (up to 49 layers) inspired by VGG and ResNets like architecture philosophy. The method used in this work starts with applying a lookup table to create a 2D dense vector where each character in fixed document length of 1024 was converted to a vector size of 16. The produced vector is then fed into 64 feature map convolutional layer of kernel size three, followed multi-convolutional a stack blocks. Each convolutional block consists of multi-temporal convolutional layers with different feature maps that depend on the layer depth and fixed kernel size of three [63]. This layer is followed by a temporal batch norm layer and a ReLU activation function. Each convolutional block is followed by a max-pooling layer of kernel size 3 and strides 2, and then double the size of convolutional feature maps of the next block. In the last block, they selected the k most important features $512 \times k$ and transformed them to a vector, which is then fed into a three fully connected layer with ReLU activation function and softmax outputs. This architecture achieved high accuracy result while using less network parameters compared to a character level CNN.

Text classification using different ways of encoding including UTF-8 bytes, characters, words, Romanized characters and Romanized words were studied in [64] using fourteen large-scale datasets with four languages (Chinese, English, Japanese and Korean). This article introduced 473-benchmarked models, including convolutional networks, linear models and fast-Text [65]. For convolutional networks, it was compared between encoding mechanisms using character glyph images, one-hot vector, and embedding. It was noticed that is a byte-level UTF-8 one-hot encoding produced a competitive results with convolutional networks. This indicates the ability of a CNN to understand text from scratch with low-level representation.

### 3.1.2 Arabic Dialect Identification

Arabic is one of the oldest ancient languages with spoken population of almost 300 million people across 22 Arabic countries knowing as Arab League countries [66]. Its composite of a group of language varieties with 30 varieties including its formal Modern Standard Arabic (MSA) [67], this many type of varieties related to mutual differences in geographical, structural, historical, social, or even ethnographic grounds [68].

Natural language processing of Arabic language is being one of the challenging tasks as its morphological language with many varieties. It is classified in to three main general categories based on literature as:
1. Classical Arabic (CA)
2. Modern Standard Arabic (MSA)
3. Dialectal Arabic (DA)

Classical Arabic is historical Arabic language since 1400 years ago and has a fixed orthography and grammar, which is not different from that time, and language of the Qur'an, and Hadith for Muslim word.

MSA is the only standardized and structured artificial language that it is no one native language, which stemmed from classical Arabic. It is commonly used for formal written communications, literary, and educational purposes across the Arabic speaking countries. It exhibits relatively minor variation, mainly in vocabulary, morphology, and phonological features [69]. For a more detailed account of the different linguistic aspects of MSA, see [66], [69], and [70].

Dialectal Arabic varieties used in informal communication throughout the Arabic world. These language varieties not only show considerable variation from one country to another, but also differ from one region to another within the same county [68] [71]. Arabic dialects are classified based on the geographical locations into five main varieties groups, as shown in Figure 6. It is commonly used and includes:

1. **Egyptian**: The variety spoken in Egypt, which is widely spread due to the historical impact of Egyptian media

2. **Gulf**: A variety spoken primarily in Saudi Arabia, UAE, Kuwait and Qatar
3. **Iraqi**: The variety spoken by the people of Iraq
4. **Levantine**: The variety spoken primarily by the Levant (i.e., people of Syria, Lebanon, and Palestine)
5. **Maghrebi**: The variety spoken by people of North Africa, excluding Egypt

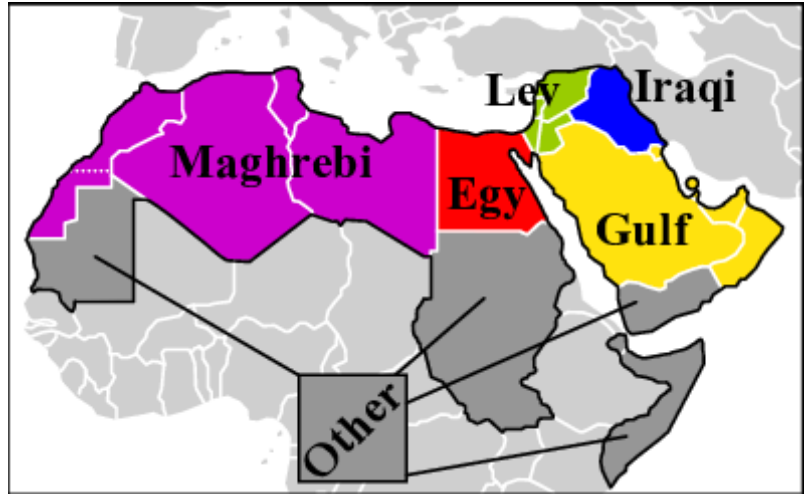

**Figure 6: Categorization of Arabic dialects** [72]

Dialectal Arabic texts considerably deviate from the formal MSA while using the same MSA rules in term of phonology, morphology, and syntax in daily spoken communications [73]. It is worth noting that the Arabic dialects have become increasingly apparent as the language of informal communication on the web in blogs, forums, and social media networks …etc. The non-standard spelling, poor quality, and common vocabulary among the different dialects pose more complexities to the identification problem of the language dialectal varieties.

The discrimination between DA and MSA is rather complex because of their interrelatedness; Arabic speakers tend to use both of them in different purposes as the situation demands in their day-to-day life. While MSA is an established standard among educated Arabic speakers, in contrast DA does not follow any standards, and is used in every day informal communication

Work done in [74] proposed a traditional supervised method using Naïve Bayes Classifier for binary classification sentence-level MSA-EGY categorization, using a subset of the AOC dataset (12,160 MSA and 11,274 Egyptian Dialect). The authors study the effect of different type of pre-processing included language modeling, Morphological

Analyzer, Orthography Normalized, and Perplexity based feature on classifier performance. They report 85.5% accuracy using 10-fold cross-validation, compared to the 80.9% accuracy reported by [72].

The work presented in [75] introduced also the binary MSA-EGY classification task using Random forest ensemble classifier. The authors exploit Twitter data collected a dataset of 880K tweets for training, while testing on 700 tweets they labeled for the task. Their studied a range of lexical and morphological features and reporting accuracy of 94.6% which representing a 10% absolute gain over models trained with n-grams only.

Author of paper [76] focus on the multi-classification task using the AOC categories (MSA, EGY, GLF,LEV). He introduced an ensemble of weakly, strong, and semi-supervised learning methods exploiting 165M data points from Facebook posts in addition to improved classification accuracy using a simple word-level n-gram model trained on the manually annotated portion of AOC as well as un-annotated Facebook data. He achieved an accuracy of 87.8% using testing dataset with size of 9k which representing 10% of the manually annotated AOC dataset.

The paper [77] introduced the idea of deep learning in Multi-classification of Arabic dialects (LEV-GLF-EGY) by implemented Four different deep neural network models to examine their effectiveness by using subset of AOC dataset with size of 33K. The authors report an accuracy of 71.4%, 68%, 71.1%, and 70.90% using LSTM, CNN, CLSTM, and BLSTM respectively.

The scholar in [78] provided a benchmarked dataset for Arabic dialects identification extracted from AOC dataset, and consist of 108,174 distributed among 4 categories (MSA, EGY, GLF,LEV) and splitting into 86,541 for training, 10,821 for validating, and 10,812 for testing. The authors used this dataset to perform 25 experiments in which 7 of them using traditional machine learning classifier, and 18 using deep learning models (CNN-CLSTM-LSTM-BiLSTM-BiGRU- Attention-BiLSTM). The author tested three different embedded settings (Random embeddings- AOC-based embeddings- Twitter-City embeddings). The highest accuracy achieved for Four-way identification was 82.45% on testing & 83.49% on validation when used Attention BiLSTM with Twitter-City embeddings.

## 3.2 Related Work Summary Table

**Table 1: Related Work Summary for English Text Classification**

| ID | Algorithm | Input Feature Vector | Dataset | Result |
|---|---|---|---|---|
| X. Zhang, J. Zhao, Y. LeCun, 2015 [10] | CNN (6 Convolutional Layers) | 70 Characters Quantization Using 1-of-m Encoding "one-hot encoding" with fixed document length (Large=1024 or small=256) | Different Dataset size from 120K-3.6M (AG's news – Sogou news – DBPedia – Yelp Review – Yahoo! Answers – Amazon Review) | Best Accuracy using ngrams TFIDF for "first 3 dataset" (92.4%-97.2%-98.7%) respectively. Using ngrams for "Yelp P." 95.6%. Best accuracy for The remaining dataset using Char-CNN (62.0%-71.2%-59.5%-94.5%) respectively. As training dataset increase the accuracy of CNN increase. |
| Joseph D. Prusa, Taghi M. Khoshgoftaar, 2016 [55] | CNN (3 Convolutional Layers) | Character encoding using "UTF-8" with fixed document length 182 | 100,000 (50K positive, 50K negative) randomly sampling from tweet sentiment140 | Increase the accuracy by almost 3.3% when using (1-of-8 encoding) compare to 70 character quantization (1-of-70 encoding) |
| Yijun Xiao and Kyunghyun Cho, 2016 [56] | Hybrid model using CNN and Recurrent layers | 96 character representation of a sentence with fixed length of 1014. Each character with 8 embedding dimension. Total input feature of (1014*8) | Different Dataset size from 120K-3.6M (AG's news – Sogou news – DBPedia – Yelp Review – Yahoo! Answers – Amazon Review) | Error rate 1.43% - 40.77% depends on Dataset. This model In general, achieve higher accuracy compare to Zhang et al [10] in 5 dataset results from total 8 dataset results |

| ID | Algorithm | Input Feature Vector | Dataset | Result |
| --- | --- | --- | --- | --- |
| A. K. Uysal, Y. L. Murphey, 2017 [58] | CNN (1 Convolutional Layers) | One hot representation, Semantic word embedding (Glove), Fine-tuned Glove. | IMDB movie review, Sentiment140, Nine public sentiment, Multi-domain | CNN first highest results with "one-hot representing" & second result with "Fine-tuned Glove" and worst one compare to other two method is Glove without fine-tuned. |
| X. Ma, R. Jin, J. Y. Paik, T. S. Chung, 2017 [62] | CNN (1 Convolutional Layers) | De-noise based word2vec; It aims at reducing the vocabulary size of large dataset by filtering out misspelled or uncommonly used words. | 560K DBPedia dataset | Dn-CNN model is less error rate 1.15% compare to "CNN word model of 1.49%" & "char-CNN model of 1.95%." |
| A. Conneau, H. Schwenk, Y. Le Cun 2017 [18] | Very Deep CNN (Up to 29 Convolutional layers) | 69-character representation of a sentence with fixed length of 1014. Each character with 16 embedding dimension. Total input feature of (1014*16) | Different Dataset size from 120K-3.6M (AG's news – Sogou news – DBPedia – Yelp Review – Yahoo! Answers – Amazon Review) | Error rate 1.29% - 37% depends on Dataset. Get closer to the accuracy of state art (n-grams TF-IDF) for small data sets and more significant accuracy result on large data sets. In general, achieve higher accuracy compare to Zhang et al [10]. |
| X. Zhang, Y. LeCun, 2017 [64] | Multi CNN architecture depends on encoding used. | Different text encoding levels are studied, including "bytes", "characters", "words", "Romanized characters"," Romanized words". | 14 datasets from 4 languages including (Chinese, English, Japanese and Korean) | Best encoding mechanism for convolutional networks is byte-level one-hot encoding "UTF-8." |

**Table 2: Related Work Summary for Arabic Dialect Identification**

| Paper ID | Algorithm | Dataset | Result |
| --- | --- | --- | --- |
| Elfardy and Diab, 2013 [74] | Binary Classification of MSA-EDA Using Naïve Bayes Classifier with Token Feature (LM-Morphological Analyzer-Orthography Normalized) & Perplexity based feature | Subset of Arabic Online Commentary (AOC) Dataset 12,160 MSA sentences and 11,274 of Egyptian news articles Comments | 85.50% |
| Darwish et al., 2014 [75] | Binary Classification of MSA-EDA Using Random forest ensemble classifier with explore a range of Dialectal Egyptian lexical and morphological features and report a 10% absolute gain over models trained with n-grams only | 880K tweets for train, and 700 for testing | 94.60% |
| Huang, 2015 [76] | Multi-classification (MSA-LEV-GLF-EGY) Using combination of weakly & strong & semi-supervised n-gram based classifier | AOC Dataset Using 9K for testing | 87.8% on 10% of the manually annotated AOC dataset. |
| Lulu and Elnagar, 2018 [77] | Multi-classification (LEV-GLF-EGY) Using LSTM – BLSTM – CNN – CLSTM | 33K AOC Dataset | 71.4% LSTM, 68% CNN, 71.1% CLSTM, 70.90% BLSTM |
| Elaraby and Abdul-Mageed, 2018 [78] | Multi-classification (MSA-LEV-GLF-EGY) Using CNN-CLSTM-LSTM-BiLSTM-BiGRU-Attention BiLSTM With three different Embeddings | 108K AOC Dataset | The Highest accuracy of 82.45% on Testing & 83.49% on Development Using Attention BiLSTM |

## 3.3 Discussion & Conclusion

One of the major challenges in text classification and NLP applications is how to encode the input text. The most commonly used methods are shown below in Figure 7. Some CNN models, deal with input text as a stream of characters. Each English alphabetic character inside the document of length l is encoding in one-hot vector of characters (one-of-m) with one value in this character position inside the vector and zeros values for other, where m size is equivalent to alphabet vocabulary size [10]. A similar method introduced for handling the character input stream using 256 Unicode characters is found in UTF-8, in which each character can be represented in a binary vector of size 8, equivalent to its Unicode value. The drawbacks of those models are the language dependent in which many variables need to be re-engineered in case of language changes in addition to its inability to handle a large character vocabulary language like Chinese.

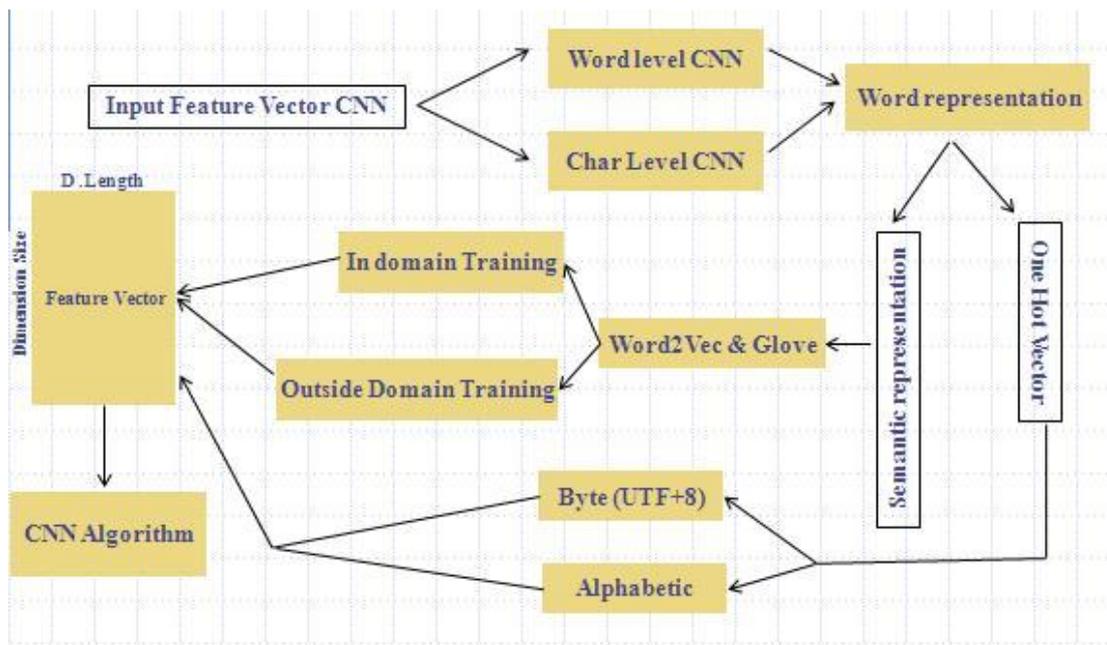

**Figure 7: Summarized on Text Encoding Methods for CNN**

Other models deal with input text as a stream of words in which each word is encoded using one-hot vector of words. The problem with the one-hot representation is the dimensionality of the word vectors. For example, in a dataset with 10K words vocabulary, each word in the text will be represented by a vector of size 10K.

A word embedding methods (i.e. Word2Vec [11] or Global vectors [79]) are commonly used for encoding the input text data which feeding into DNNs. This type of encoding is effective in dealing with the problem of the dimensionality by representing the word into lower dimensional vector space in addition to its ability to capture many semantic relationships between the words they represent. Although pre-trained word2vec embedding is intensively available for English, it is costly to obtain and train for some languages. Moreover, it does not allow benefiting from the powerful of DNN in automatically feature extraction from raw data while semantic representation is used.

# Chapter 4: The Proposed Model

The main components of the proposed model as shown below in Figure 8 have two-step process: **First**, creating the text-encoding feature vector for feeding into CNN input layer from raw text data using BUNOW and BUNOC methods, which are described in detail in Section 4.1 for text classification and Arabic dialects identification. **Second**, using CNN Architecture models described in Section 4.2 for text classification and Arabic dialects identification.

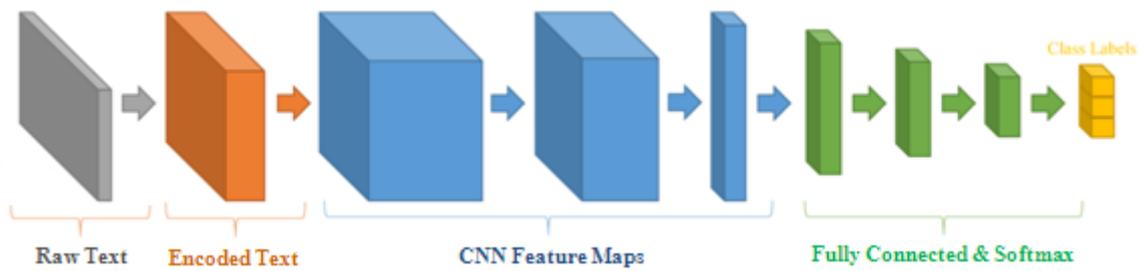

**Figure 8: Model Architecture**

## 4.1 Text Encoding Methods

The main components of the proposed text encoding methods are shown below in Figure 9.

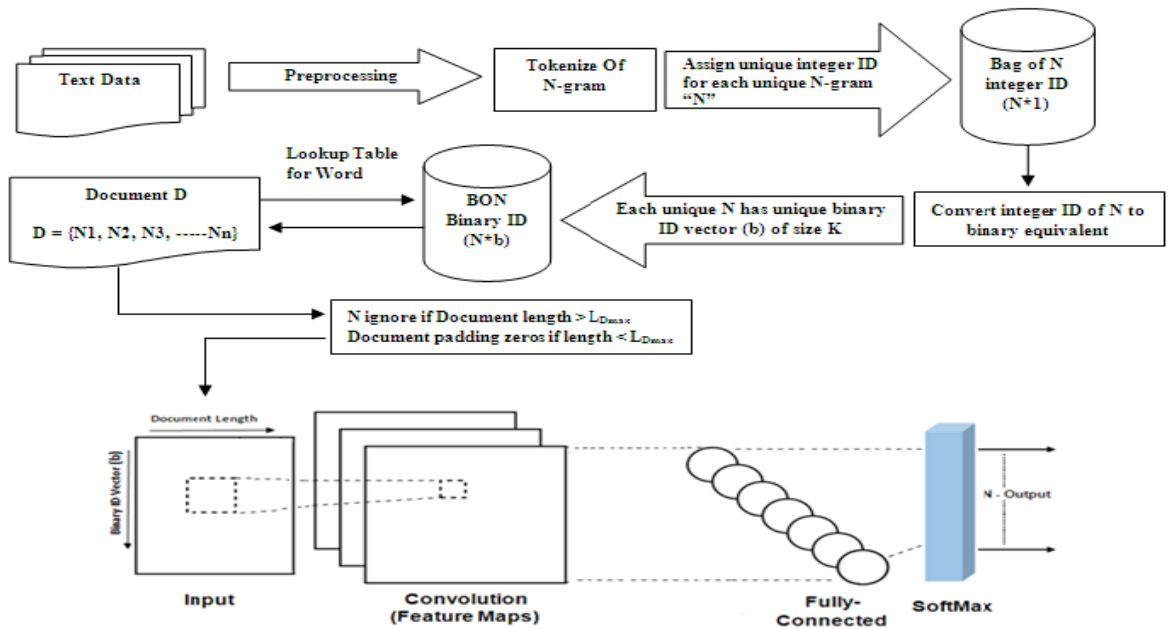

**Figure 9: BUNOW & BUNOC Encoding Methods**

### 4.1.1 Binary Unique Number of Word Method (BUNOW)

The rationale behind the BUNOW method is to convert the unique serial integer ID ($IDw_i$) given to each unique word ($w_i$) in the training corpus (T) to its binary equivalent number ($Bw_i$), and represent it as a fixed k-dimensional vector where its dimension equal to ($2^k$ = Total $IDw_i$ in T). The following steps are applied to BUNOW Model:

1) Create vocabulary of unique words from training dataset.

2) Assign unique serial integer ID for each word.

3) Convert the serial integer ID for each word to its equivalent binary value, and represent it in a fixed dimensional vector (b) of size k, where ($2^k$ = vocabulary size).

4) Set the maximum document length ($LD_{max}$) based on the document with maximum number of words ($W_{max}$) in the training dataset.

The input feature vector (IV) is created by concatenating a binary vector of each word ($bw_i$) that exists in a document (D), such that:

$$IVD = bw_1 + bw_2 + bw_3 + bw_4 + \ldots + bw_{max} \qquad \ldots\ldots\ldots\ldots\ldots\ldots\ldots\ldots\ldots (1)$$

Where (+) is the concatenation operator.

Any word that exceeds the document length ($LD_{max}$) is ignored, and any document less than ($LD_{max}$) are padded with all-zero vectors. In the testing phase, instead of using zeroes for an unseen word, we gave each unique unseen word an equivalent binary value equal to its subsequent integer ID in vocabulary (V) and then updates V under the following constrain: the representation of the binary value of an unseen word would not exceed the initialized fixed dimensional vector (b) of the word, if exceeded then a vector of zeros for this word will be used.

### 4.1.2 Binary Unique Number of Character Method (BUNOC)

The rationale behind the BUNOC method is to convert the unique serial integer ID ($IDw_i$) given to each N-gram character ($NC_i$) in the training corpus (T) to its binary equivalent number ($BNC_i$), and represent it as a fixed k-dimensional vector where its dimension equal to ($2^k$ = Total $IDNC_i$ in T). the following steps are applied to BUNOC Model:

1) Create vocabulary of unique N-gram character from training dataset.

2) Assign unique serial integer ID for each N-gram character.

3) Convert the serial integer ID for each N-gram character to its equivalent binary value, and represent it in a fixed dimensional vector (b) of size k, where ($2^k$ = vocabulary size).

4) Set the maximum document length ($LD_{max}$) based on the document with maximum number of N-gram character ($NC_{max}$) in the training dataset.

The input feature vector (IV) is created by concatenating a binary vector of each N-gram character ($bnc_i$) that exists in a document (D), such that:

$$IVD = bnc_1 + bnc_2 + bnc_3 + bnc_4 + \ldots + bnc_{max} \quad \ldots\ldots\ldots\ldots\ldots\ldots\ldots\ldots (2)$$

Where (+) is the concatenation operator.

Any N-gram character that exceeds the document length ($LD_{max}$) is ignored, and any document less than ($LD_{max}$) are padded with all-zero vectors. In the testing phase, instead of using zeroes for an unseen N-gram character, we gave each unique unseen N-character an equivalent binary value equal to its subsequent integer ID in vocabulary (V) and then updates V under the following constrain: the representation of the binary value of an unseen N-gram character would not exceed the initialized fixed dimensional vector (b) of the N-gram character, if exceeded then a vector of zeros for this N-gram character will be used.

## 4.2 CNN Architecture

The provided model in this paper is partially inspired by Alex-Net [30], a large deep CNN model that competed in 2012 challenge (Image-Net Large Scale Visual Recognition) and classified as a top-5 winner, where achieved an accuracy rate of 84.7% compared to 73.8% achieved by the second-best entry. Their architecture is characterized by using a stack of convolutional layers with max pooling, and ReLU activation function, three fully connected layers with dropout 0.5, and Softmax classifier in the last fully connected layer [80]. A similar architecture was used by [10] for text classification where the researchers used six 1D temporal convolutional layers of kernels size 3, and 7 with convolutional feature maps of 256 for small feature, while using 1024 for large feature followed by simple Max-pooling layers, and three fully connected layers.

### 4.2.1 English Text Classification Model

The architecture of the proposed CNN is summarized in Figure 10. It consists of eight deep spatial convolutional layers, which simply computes a 2-D convolution and three fully connected layers. The architecture is divided into three blocks; the first block consists of four convolutional layer with feature maps 200, stride (1, 1), and ReLU activation function for each. Vertical convolutional is applied over input vector while ignoring horizontal convolution by using kernel size ((one, 7), (one, 7), (one, 3), (one, 3)) respectively with valid padding to learning the structure of each word separately from its binary k-dimensional vector. This allows the network to understand the words based on their position inside document while ignoring the relation between them.

The second block consists of four convolutional layer with stride (1, 1), ReLU activation function for each layer, and feature maps (150, 100, 50, 25), respectively. Horizontal convolutional is then applied over output feature maps of the previous block by using kernel size ((three, 1), (three, 1), (seven, 1), (seven, 1)) respectively with valid padding in order to let the network understand the relation between words inside the document and distinguish between them.

Finally, three fully connected layers of 2048, 1024, and 4 are used after the last convolutional layer in which the last fully connected layer connect to softmax classifier to classify the document based on four class category. In order to regularize the network, and prevent over-fitting, we use five dropout of 0.5 after the fourth, $6^{th}$, $8^{th}$, $9^{th}$, and $10^{th}$ layer, respectively [81].

A max-pooling layer is commonly used to reduce a feature map size of output convolutional layer by selecting the maximum feature value before connecting it to another convolutional layer or a fully connected layer to allow fewer weight parameters, which directly affects memory and computational consumption. It was found that it is not useful to use the pooling layer with low-level convolutional feature map of size 165 each as this leads to loss of information. The alternative was to reduce the convolutional feature maps number from 200 to 25 in the last convolutional layer. This helps to achieve high accuracy without loss of information compared to the current state of the art.

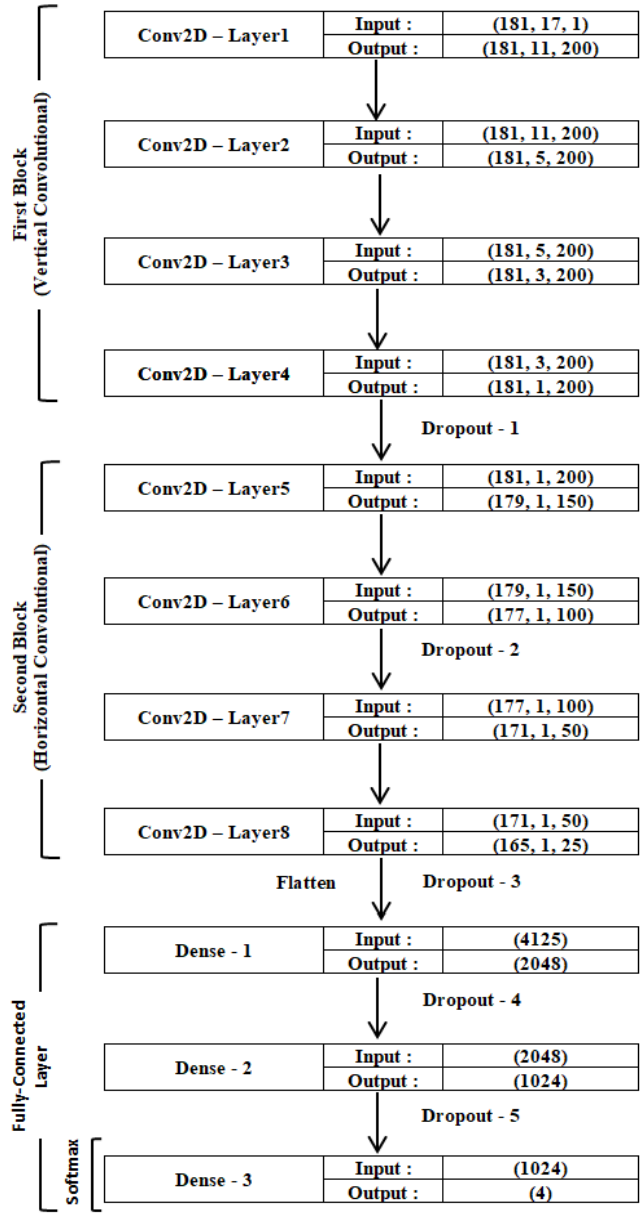

Figure 10: CNN BUNOW Model for English Text Classification

## 4.2.2 Arabic Dialect Identification Model

The English text CNN architecture is used with minor adjustment to fit a variant of Arabic text input dimension size. The architectures of the proposed CNNs for BUNOW, Bi-BUNOC, and Tri-BUNOC are summarized in Figure 11, Figure 12, and Figure 13, respectively.

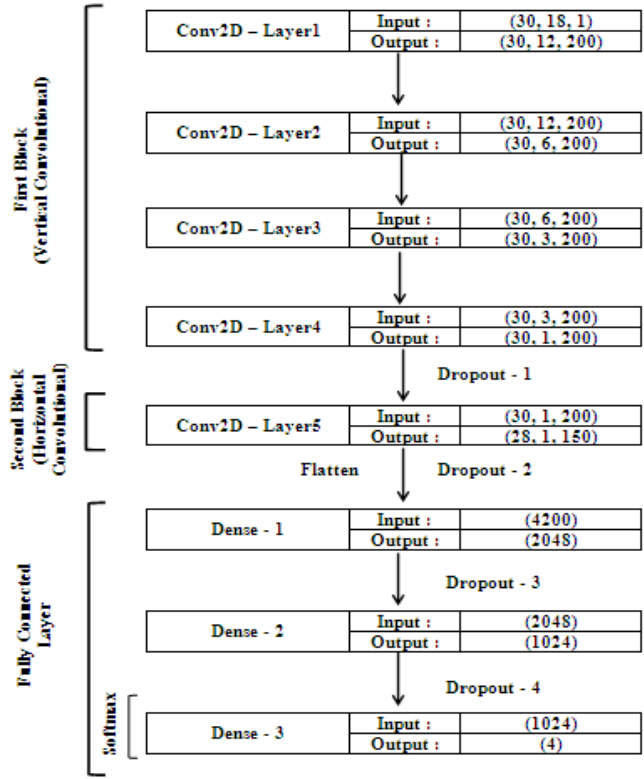

**Figure 11: Arabic Dialect Identification BUNOW Model**

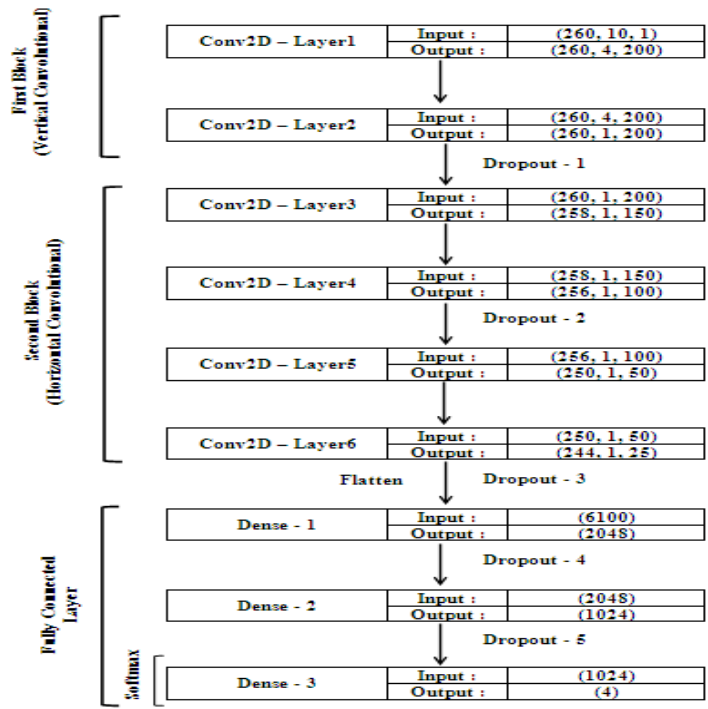

**Figure 12: Arabic Dialect Identification Bi-BUNOC Model**

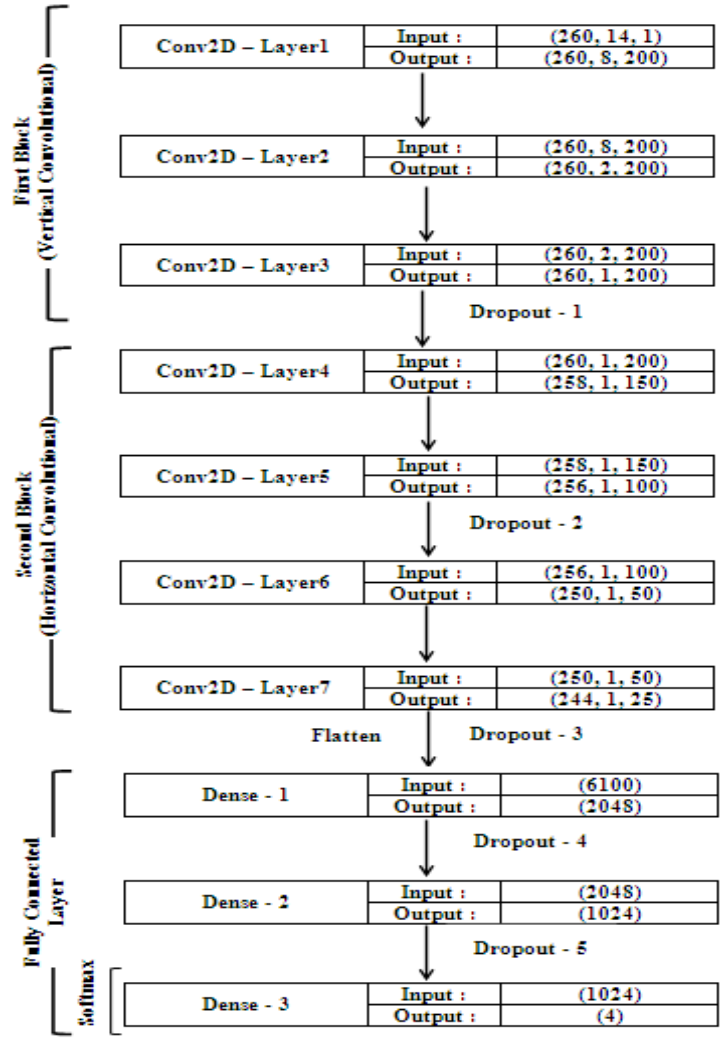

**Figure 13: Arabic Dialect Identification Tri-BUNOC Model**

BUNOW used word for representing the input text data, it consists of five deep spatial convolutional layers, which simply computes a 2-D convolution, and three fully connected layers. The architecture is divided into three blocks; the first block consists of four convolutional layer with feature maps 200, stride (1, 1), and ReLU activation function for each. Vertical convolutional is applied over input vector while ignoring horizontal convolution by using kernel size ((one, 7), (one, 7), (one, 4), (one, 3)) respectively with valid padding to learning the structure of each word separately from its binary k-dimensional vector. This allows the network to understand the words and be able to distinguish between them. The second block consists of one convolutional layer with stride (1, 1), ReLU activation function, and feature map of size 150. Horizontal convolutional is then applied over output feature map of the previous block by using kernel size ((three, 1), with valid padding in order to let the network understand the semantic and the relation between words inside the document and distinguish between them. Finally, three fully connected layers of 2048, 1024, and 4 are used after the last convolutional layer in which the last fully connected layer connect to softmax classifier to classify the document based on four class category. In order to regularize the network, and prevent over-fitting, we use four dropout of 0.5 after the fourth, $5^{th}$, $6^{th}$, $7^{th}$, and $8^{th}$ layer, respectively [81].

Bi-BUNOC used Bigram character for representing the input text data, it consists of six deep spatial convolutional layers, which simply computes a 2-D convolution and three fully connected layers. The architecture is divided into three blocks: the first block consists of two convolutional layer with feature maps 200, stride (1, 1), and ReLU activation function for each. Vertical convolutional is applied over input vector while ignoring horizontal convolution by using kernel size ((one, 7), (one, 4)) respectively with valid padding to learning the structure of each word separately from its binary k-dimensional vector. This allows the network to understand the words based on their position inside document while ignoring the relation between them. The second block consists of four convolutional layer with stride (1, 1), ReLU activation function for each layer, and feature maps (150, 100, 50, 25), respectively. Horizontal convolutional is then applied over output feature maps of the previous block by using kernel size ((three, 1), (three, 1), (seven, 1), (seven, 1)) respectively with valid padding in order to let the network understand the relation between words inside the document and distinguish between them. Finally, three fully connected layers of 2048, 1024, and 4 are used after the last convolutional layer in which the last fully connected layer connect to softmax classifier to classify the document based on four class category. In order to regularize the network, and prevent over-fitting, we use five dropout of 0.5 after the $2^{nd}$, $4^{th}$, $6^{th}$, $7^{th}$, and $8^{th}$ layer, respectively [81].

Tri-BUNOC used Trigram character for representing the input text data, it consists of seven deep spatial convolutional layers, which simply computes a 2-D convolution, and three fully connected layers. The architecture is divided into three blocks; the first block consists of three convolutional layer with feature maps 200, stride (1, 1), and ReLU activation function for each. Vertical convolutional is applied over input vector while ignoring horizontal convolution by using kernel size ((one, 7), (one, 7), (one, 2)) respectively with valid padding to learning the structure of each word separately from its binary k-dimensional vector. This allows the network to understand the words based on their position inside document while ignoring the relation between them. The second block consists of four convolutional layer with stride (1, 1), ReLU activation function for each layer, and feature maps (150, 100, 50, 25), respectively. Horizontal convolutional is then applied over output feature maps of the previous block by using kernel size ((three, 1), (three, 1), (seven, 1), (seven, 1)) respectively with valid padding in order to let the network understand the relation between words inside the document and distinguish between them. Finally, three fully connected layers of 2048, 1024, and 4 are used after the last convolutional layer in which the last fully connected layer connect to softmax classifier to classify the document based on four class category. In order to regularize the network, and prevent over-fitting, we use five dropout of 0.5 after the $3^{rd}$, $5^{th}$, $7^{th}$, $8^{th}$, and $9^{th}$ layer, respectively [81].

A max-pooling layer is commonly used to reduce a feature map size of output convolutional layer by selecting the maximum feature value before connecting it to another convolutional layer or a fully connected layer to allow fewer weight parameters, which directly affects memory and computational consumption. It was found that it is not useful to use the pooling layer with low-level convolutional feature map of size as this leads to loss of information. This helps to achieve high accuracy without loss of information compared to the current state of the art.

# Chapter 5: Experiment Scenarios & Results

## 5.1 Datasets

Different datasets were used for English and Arabic text classification as described in the subsections below.

### 5.1.1 English Text Classification Model

The dataset used is from AG's News that can be obtained from news articles on the web. It approximately contains of 496K articles extracted from more than 2000 news sources. The selected categories from this corpus are the four largest ones (Sports, Sci/Tech, Business, and World). The title and description fields are used from these categories to build this dataset, as per samples shown below in Table 3. From each category, 30,000 samples were randomly chosen for training and 1,900 for testing [10]. Samples size is about 120,000 for training and 7,600 for testing. The dataset statistic and classes label is shown below in Table 4 and Table 5 respectively.

**Table 3: Samples of AG's News Dataset – English Text Classification**

| Class # | Title | Description |
|---|---|---|
| 1 | On front line of AIDS in Russia. | An industrial city northwest of Moscow struggles as AIDS hits a broader population. |
| 2 | Giddy Phelps Touches Gold for First Time. | Michael Phelps won the gold medal in the 400 individual medley and set a world record in a time of 4 minutes 8.26 seconds. |
| 3 | Fears for T N pension after talks. | Unions representing workers at Turner Newall say they are 'disappointed' after talks with stricken parent firm Federal Mogul. |
| 4 | IBM Chips May Someday Heal Themselves. | New technology applies electrical fuses to help identify and repair faults. |

**Table 4: AG's News Statistic – English Dataset**

| Class # | Language | Training # | Testing # | Unique Words # | Max Doc Length # |
|---|---|---|---|---|---|
| 4 | English | 120,000 | 7,600 | 70,396 | 181 |

**Table 5: AG's News Classes Label – English Dataset**

| Class # | 1 | 2 | 3 | 4 |
|---|---|---|---|---|
| Name | World | Sports | Business | Science/Technology |

## 5.1.2 Arabic Dialect Identification Model

Our work is based on the AOC dataset. AOC is composed of 3M MSA and dialectal comments [72], of which 108,173 comments are labeled via crowdsourcing. For our experiments, we used the same dataset of paper [78] which splits into 80% training (Train), 10% validation (Val), and 10% test (Test). Table 6 shows the distribution of the data among 4-way variants (MSA- EGY- GLF- LEV), where a given samples of dataset is shown below in Figure 14.

**Table 6: Distribution of AOC Dataset classes – Arabic Dialects**

| Variety | MSA | EGY | GLF | LEV | All |
|---|---|---|---|---|---|
| Train | 50,845 | 10,022 | 16,593 | 9,081 | 86,541 |
| Val | 6,357 | 1,253 | 2,075 | 1,136 | 10,821 |
| Test | 6,353 | 1,252 | 2,073 | 1,133 | 10,812 |

| Variety | Example |
|---|---|
| MSA | شدوا الهمة يا منتخبنا ، أنتم لا تسيرون وحدكم (1)<br>Go, go, our team; you've our passionate support.<br>بصراحة لقد عجزت عن الكلام فلا الجهل يبرر هذه الفعلة ولا حتى اي سبب اخر (2)<br>Frankly, I'm speechless. Neither ignorance nor any other reason justify this action. |
| EGY | بتقولوا ان اعلامنا وفضائيتنا فجروا الازمة وكانوا مجرمين في نظركم (3)<br>You say our media and satellite channels initiated the crisis and were criminals in your review.<br>قمر صناعي ولا قمر ١٤، في الشمش لو عرف القمر يلف في الجو (3)<br>Either its a satellite or a full moon [playful for "beautiful female"], it will never rotate in its orbit correctly. |
| GLF | يعطي العافية على من اقترح هذا القرار ومن ساهم في تطبيقه (5)<br>Healthy be the one who proposed this decision, and those who contributed in applying it.<br>عندنا بعض الناس ما يرضيهم شيء ولا يكفيهم ولا يبغون يشتغلون شيء (6)<br>Nothing would please nor be enough for some of these people; they don't even want to put any efforts. |
| LEV | وانا كمان لا بصلح سيارتي ولا على بالي وأبوي وأخوي بيتكفلوا فيها (7)<br>And I also won't repair my car, nor do I care. My brother and dad will take care of it.<br>لان مو معقول توصل الأمور لهلدرجة وكانت رح تصير مليون مشكلة بسبب أشاعاتهم لطلب الجامعة (8)<br>Because it isn't reasonable for things to get to that bad. There is a million problems college students have because of their rumors. |

**Figure 14: Arabic Dialect Examples**

## 5.2 Model Setting

### 5.2.1 English Text Classification Model

The following setting has been taken under consideration:
1. Processing is performed on word-level, which is the atomic representation of a text document.
2. The vocabulary size consists of 70396 unique words after preprocessing by removing (! " # $ % & ( ) * + , - . / : ; < = > ? @ [ \\ ] ^ _ ` { | } ~ \t \n), and convert all words to lower case.
3. Post padded technique used to ensure all input document has same length of 181 words.
4. For each word, a BUNOW vector length is 17.
5. Training is performed with Adam optimizer, and Categorical Cross-Entropy loss function, using a batch size of size 120, learning rate of 0.0005.
6. Neural network weights are initialized using Xavier Uniform Initializer [82], and bias initialize with zeroes.
7. One epoch took 4.3 minute. It took 60 epochs to converge where data shuffle on each epoch.
8. Implementation is done using python and deep learning library (Keras ) [83]
9. Single GeForce GTX 1060 Max-Q with GPU memory 6G DDR5 (laptop version)

The regularization of our model and the effect of epoch number are visually shown below in Figure 15 and Figure 16.

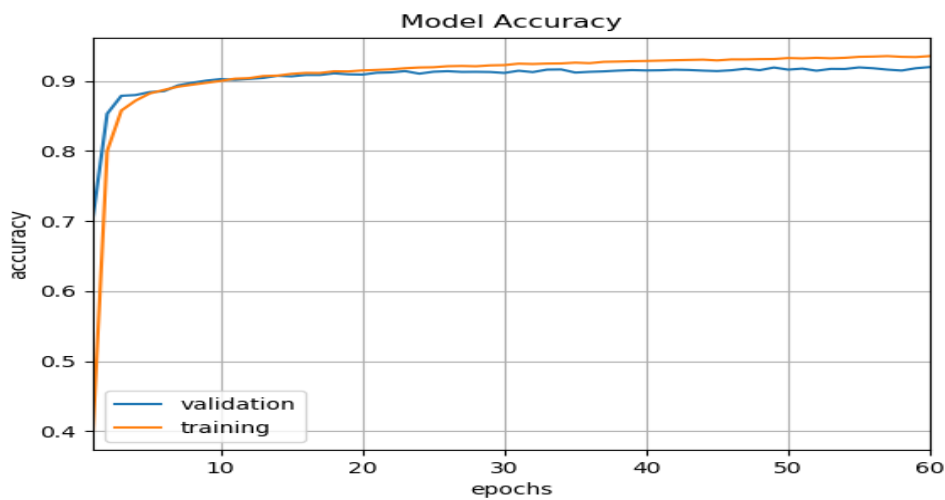

Figure 15: Training Accuracy for English Text – BUNOW Model

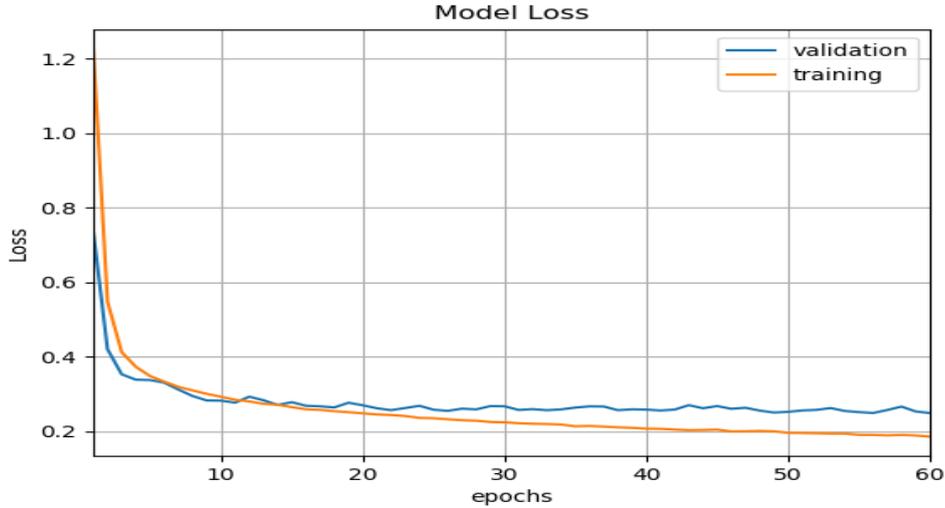

Figure 16: Training Error for English Text – BUNOW Model

### 5.2.2 Arabic Dialect Identification Model

The results from this work is compared to the results in [78], we applied the same preprocessing as following:-

1. Tokenization and normalization: We tokenize our data based on white space, excluding all non-Unicode characters. We then normalize Alif maksura to Ya, reduce all hamzated Alif to plain Alif, and remove all non-Arabic characters/words (e.g., "very", "50$").
2. All input sequences are truncated to arbitrary maximum sequence length of 30 words per comment. Comments of length < 30 are zero-padded.

The following setting of our model has been taken under consideration:

1. The vocabulary size consists of 146348 unique words for BUNOW method, 15537 for BUNOC tri-gram, and 951 for BUNOC bi-gram.
2. All input document has fixed length of 30 words for BUNOW and 260 n-gram characters for BUNOC equivalent to representing only 30 words.
3. For each word under BUNOW method, vector length is 18.
4. For each character n-gram under BUNOC method, vector length is 14, and 10 for tri-gram, and bi-gram respectively.
5. Training is performed with Adam optimizer, and Categorical Cross-Entropy loss function, using a batch size of 64, learning rate of 0.0005.

6. Neural network weights are initialized using Xavier Uniform Initializer [82], and bias initialize with zeroes.
7. One epoch took 58, 86, 144 seconds for BUNOW, Bi-BUNOC, Tri-BUNOC respectively. The data shuffle on each epoch.
8. Implementation is done using python and deep learning library (Keras) [83]
9. Single GeForce GTX 1060 Max-Q with GPU memory 6G DDR5 (laptop version)

The regularization of our models and the effect of epoch number are visually shown below from Figure 18 to Figure 22.

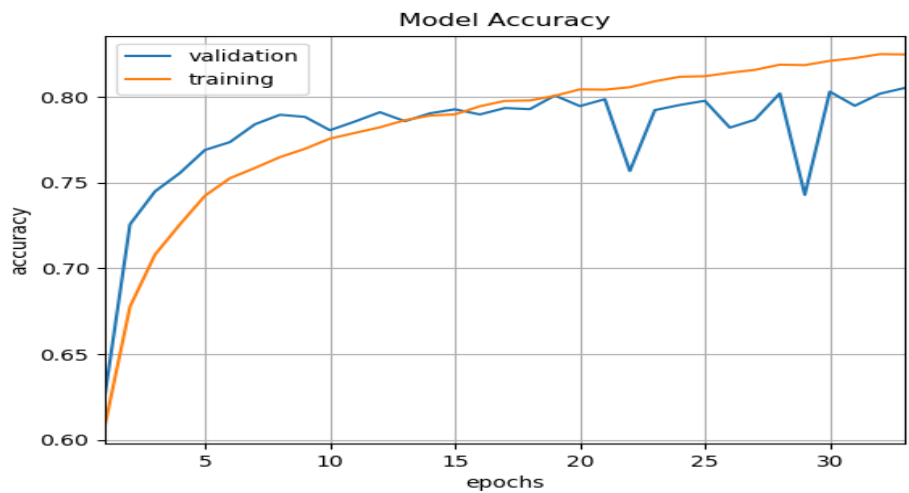

Figure 17: Training Accuracy for Arabic Dialects – BUNOW Model

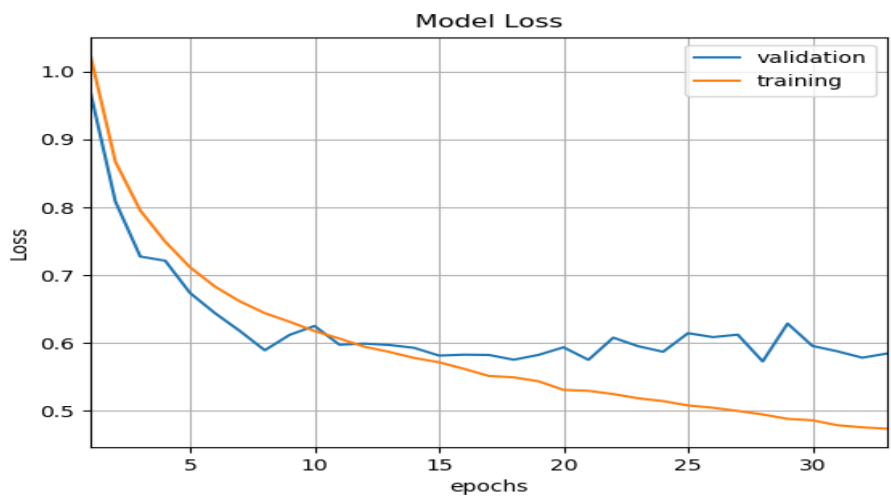

Figure 18: Training Error for Arabic Dialects – BUNOW Model

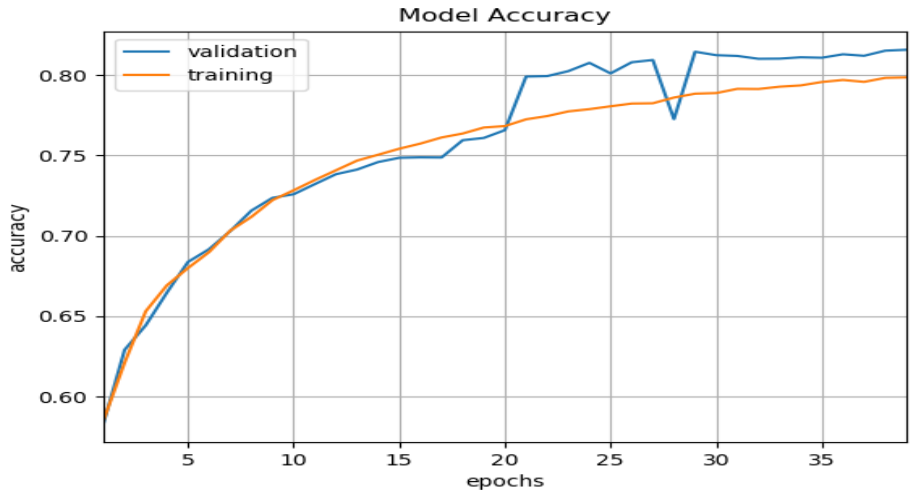

**Figure 19: Training Accuracy for Arabic Dialects – Bi-BUNOC Model**

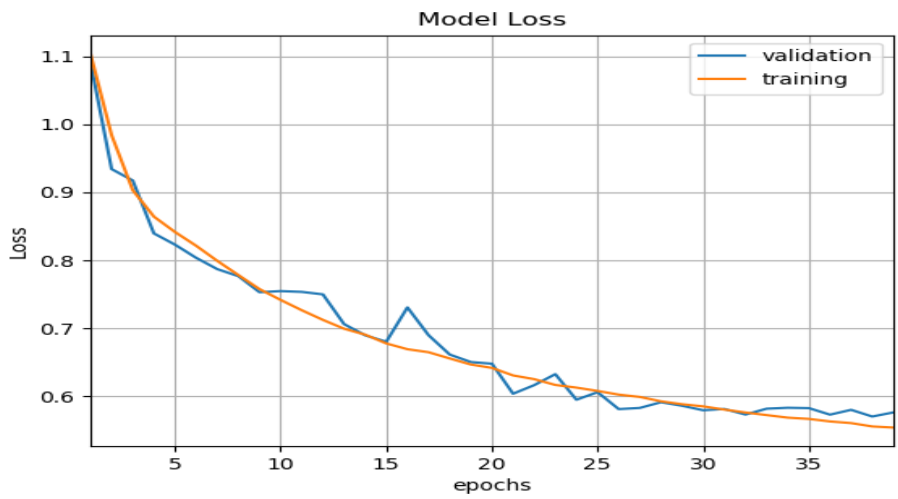

**Figure 20: Training Error for Arabic Dialects – Bi-BUNOC Model**

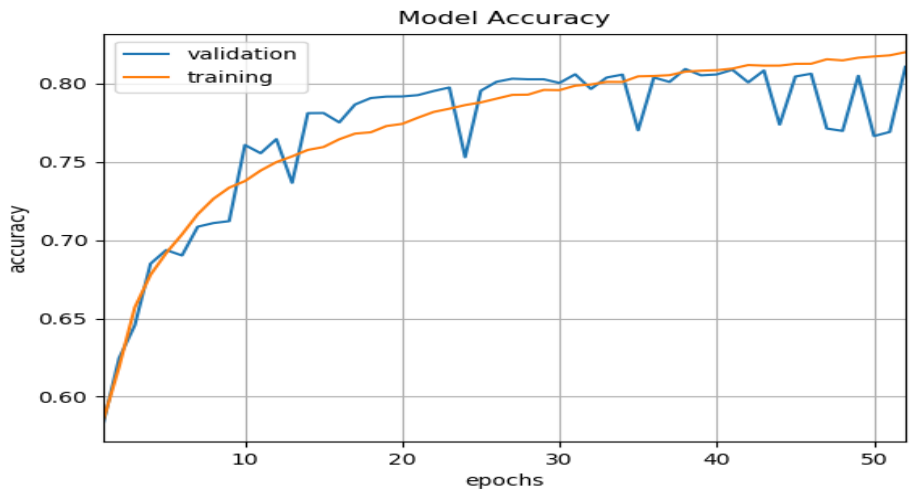

**Figure 21: Training Accuracy for Arabic Dialects – Tri-BUNOC Model**

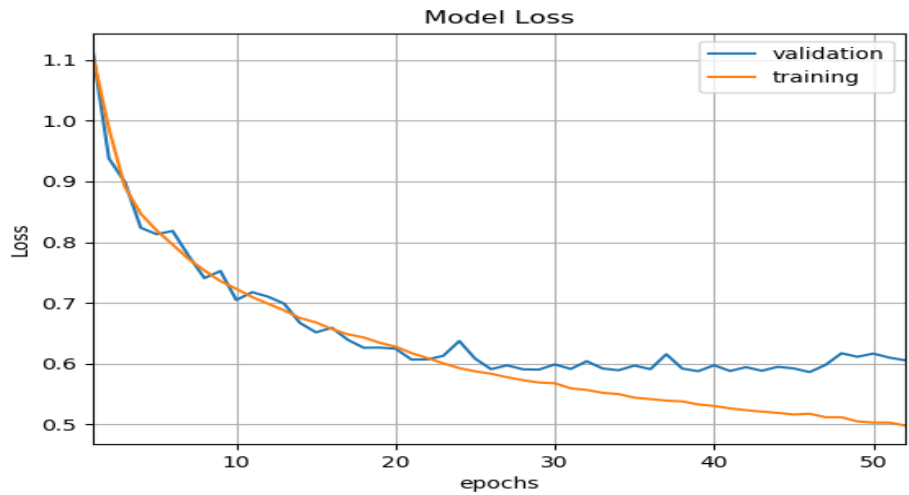

**Figure 22: Training Error for Arabic Dialects – Tri-BUNOC Model**

## 5.3 Evaluation & Results

### 5.3.1 English Text Classification Model

Our results are shown in Table 7 and confusion matrix in Figure 23. Test Error percentage was measured for each class along with overall Test Error percentage that was computed by the following equation

$$\boldsymbol{Test\ Error}\ \% = \frac{\text{Number of incorrect results}}{\text{Number of total results}} \times \boldsymbol{100} \quad \ldots\ldots\ldots\ldots\ldots\ldots\ldots\ldots\ldots\ldots\ldots\ldots\ (3)$$

**Table 7: The Proposed Model Test Error percentage – English Text Classification**

| AG's News dataset | | | | |
|---|---|---|---|---|
| Class | World | Sports | Business | Sci/Tech |
| Test Error % | 9.53 | 2.63 | 11.84 | 8.05 |
| Overall Test Error % | 8.01 | | | |
| Vocabulary# | 70396 | | | |
| OOV# | 1658 | | | |

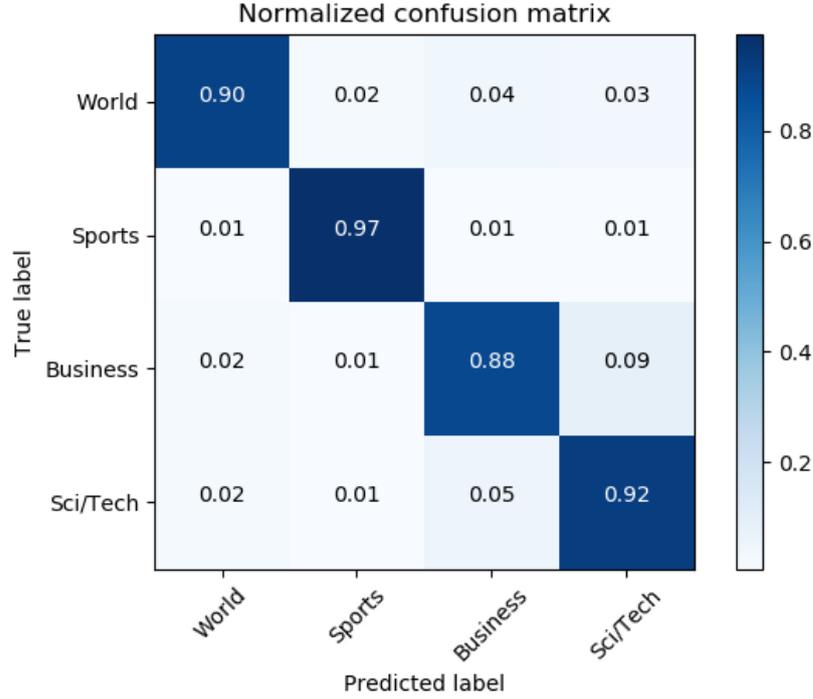

**Figure 23: The Proposed Model Confusion Matrix – English Text Classification**

Table 8 and Figure 24 visually compares the performance of the proposed model with the other CNN models results achieved from 16 experiment introduced in [10], and 9 experiment models introduced in [18], in addition to convolution recurrent network (char-CRNN) introduced in [56] using AG's News dataset. The performance of our approach compared to best of other approaches is shown in Table 9.

**Table 8: Comparison with Other Model for English Text Classification**

| Model | Test Error % |
| --- | --- |
| Our Model | 8.01 |
| Large Lookup Conv. (1) | 8.55 |
| Char-CRNN (3) | 8.64 |
| 29 Layer KMaxPooling (2) | 8.67 |
| 29 Layer MaxPooling (2) | 8.73 |
| 17 Layer MaxPooling (2) | 8.88 |
| Large Lookup Conv. Thesaurus (1) | 8.93 |
| Small Lookup Conv. Thesaurus (1) | 9.12 |
| 9 Layer MaxPooling (2) | 9.17 |
| 17 Layer Convolution (2) | 9.29 |
| 29 Layer Convolution (2) | 9.36 |
| 17 Layer KMaxPooling (2) | 9.39 |
| Large Full Conv. Thesaurus (1) | 9.51 |
| 9 Layer KMaxPooling (2) | 9.83 |
| Large Full Conv. (1) | 9.85 |
| Large Word2vec Conv. Thesaurus (1) | 9.91 |
| Large Word2vec Conv. (1) | 9.92 |
| 9 Layer Convolution (2) | 10.17 |
| Small Lookup Conv. (1) | 10.87 |
| Small Word2vec Conv. Thesaurus (1) | 10.88 |
| Small Full Conv. Thesaurus (1) | 10.89 |
| Small Word2vec Conv. (1) | 11.35 |
| Small Full Conv. (1) | 11.59 |
| Large Conv. (1) | 12.82 |
| Large Conv. Thesaurus (1) | 13.39 |
| Small Conv. Thesaurus (1) | 14.8 |
| Small Conv. (1) | 15.65 |

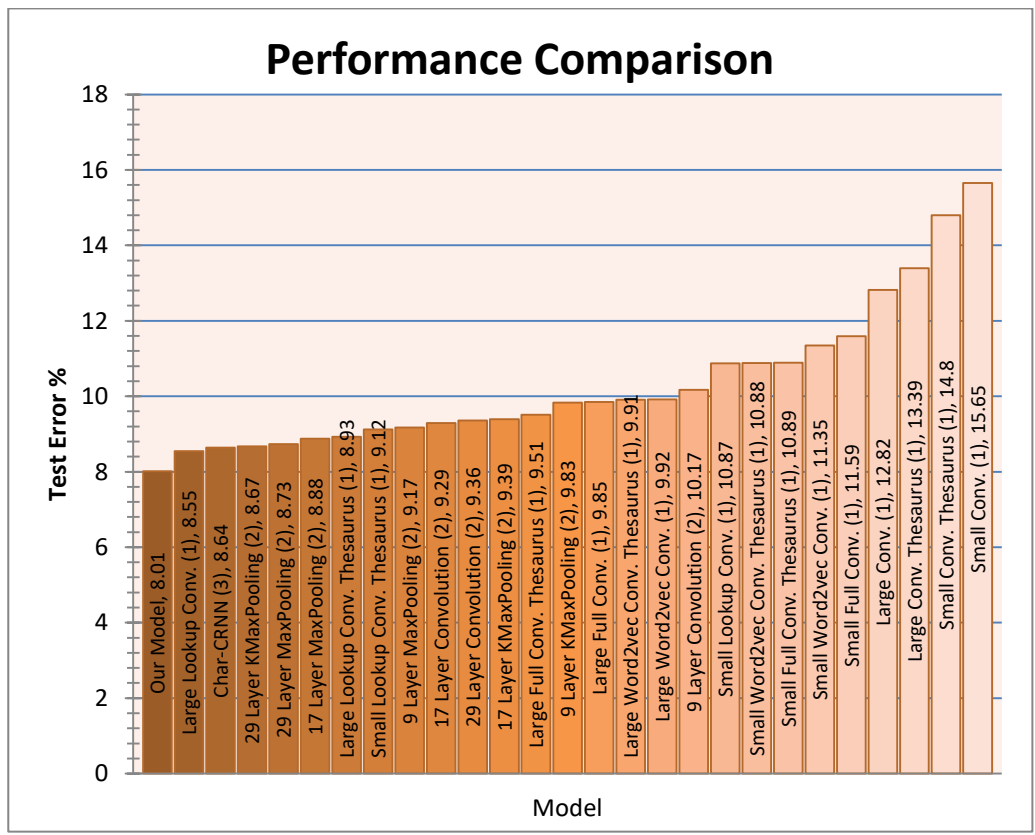

**Figure 24: Comparison Sheet for English Text Classification**

**Table 9: Comparison Result Summary for English Text Classification**

An embedding layer transforms each character to a k-dimensional vector space in order to reduce the one-hot vector size of character before feeding into CNN. Parameter# is approximated.

| Best of | Test Error % | Embedded Layer Size | CNN Input Vector Size | Parameter # in Million |
|---|---|---|---|---|
| [10] 2015 | 8.55 | N/A | 1014 x 70 | 95 |
| [56] 2016 | 8.64 | 96 x 8 | 1014 x 8 | 20 |
| [18] 2017 | 8.67 | 69 x 16 | 1014 x 16 | 17.2 |
| Proposed | 8.01 | N/A | 181 x 17 | 11.3 |

The proposed model achieved a better overall test error rate of 8.01% compared to 8.55% achieved by the best one in terms of accuracy. Note: a comparison based on individual classes was not possible as no information related to individual classes was provided in [10] [56] [18]. More importantly, the input feature vector size and number of neural network parameters in the proposed model are less than the best one by almost 62% and 34%, respectively, where the visually compares is shown in Figure 25 and Figure 26.

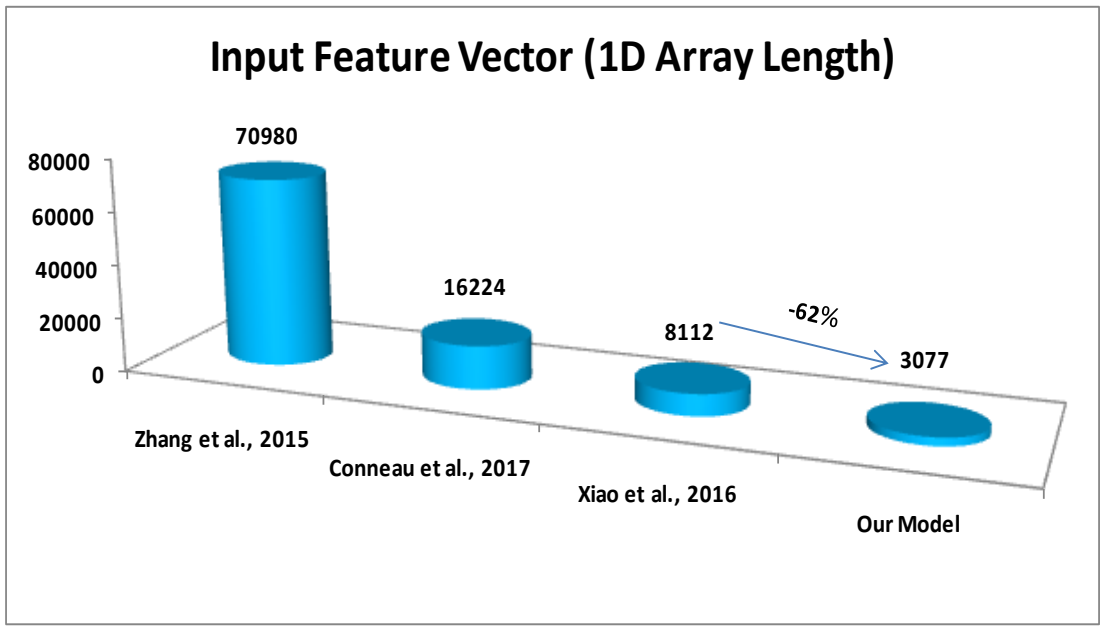

Figure 25: Input Feature Vector Compared With Other Models for English Text Classification

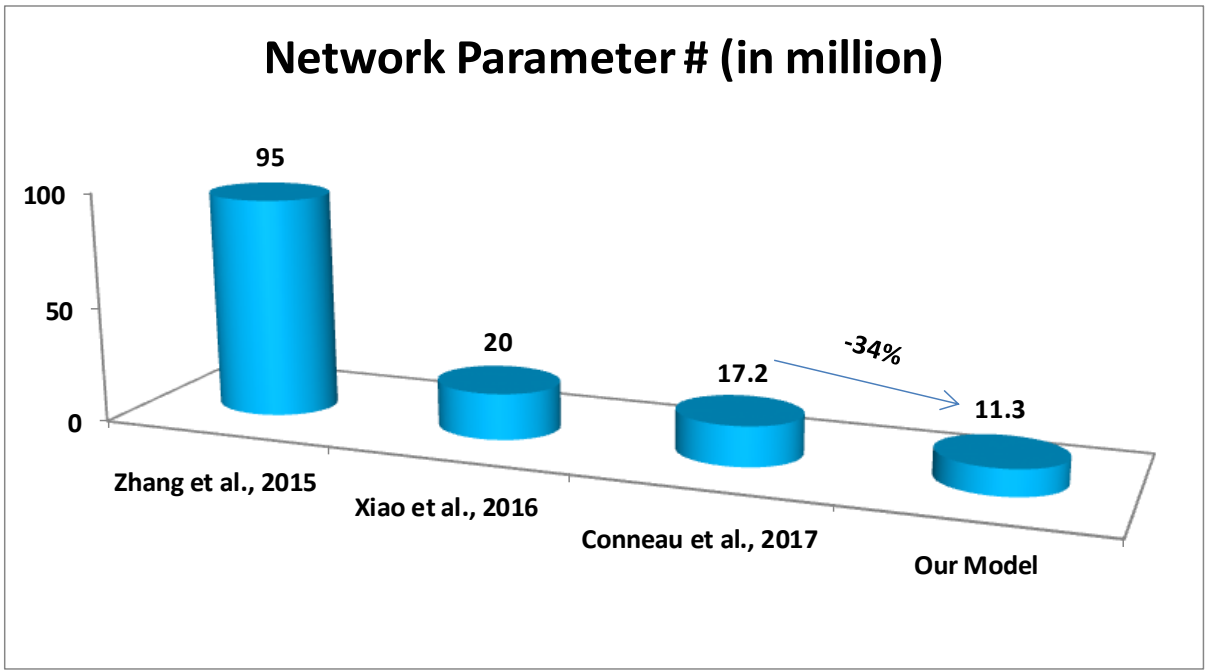

Figure 26: Network Parameter # Compared To Other Models for English Text Classification

### 5.3.2 Arabic Dialect Identification Model

Our results for validating and testing dataset are shown in Table 10 and confusion matrix of BUNOC Bi-gram in Figure 27 and Figure 28 respectively. Overall accuracy percentage was measured computed by the following equation

$$Accuracy\ \% = \frac{Number\ of\ incorrect\ results}{Number\ of\ total\ results} \times 100 \quad \dots\dots\dots\dots\dots\dots\dots\dots\dots\dots\ (3)$$

Table 10: The Proposed Model Accuracy percentage in Four-way Arabic Dialect

| Model | Validating % | Testing % | Vocabulary# | Valid-OOV# | Test-OOV# |
|---|---|---|---|---|---|
| BUNOC Bi-gram @ 39 Epoch | 81.59 | 79.76 | 951 | 9 | 6 |
| BUNOC Tri-gram @ 52 Epoch | 81.07 | 79.47 | 15537 | 251 | 215 |
| BUNOW @ 33 Epoch | 80.52 | 78.41 | 146348 | 7807 | 8284 |
| BUNOW @ 10 Epoch | 78.62 | 77.00 | 146348 | 7807 | 8284 |

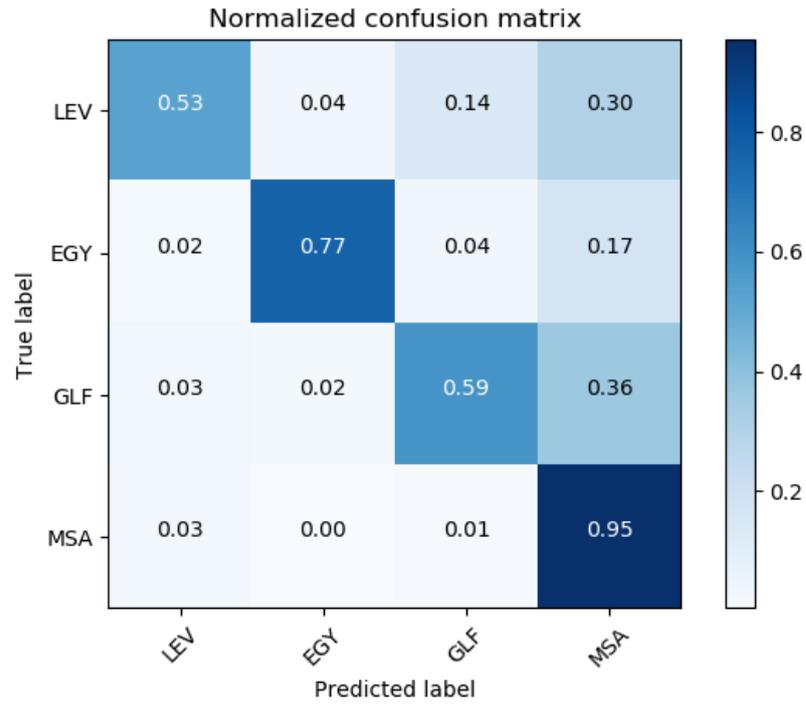

**Figure 27: BUNOC Bi-gram Model Confusion Matrix in Validation Arabic Dialect**

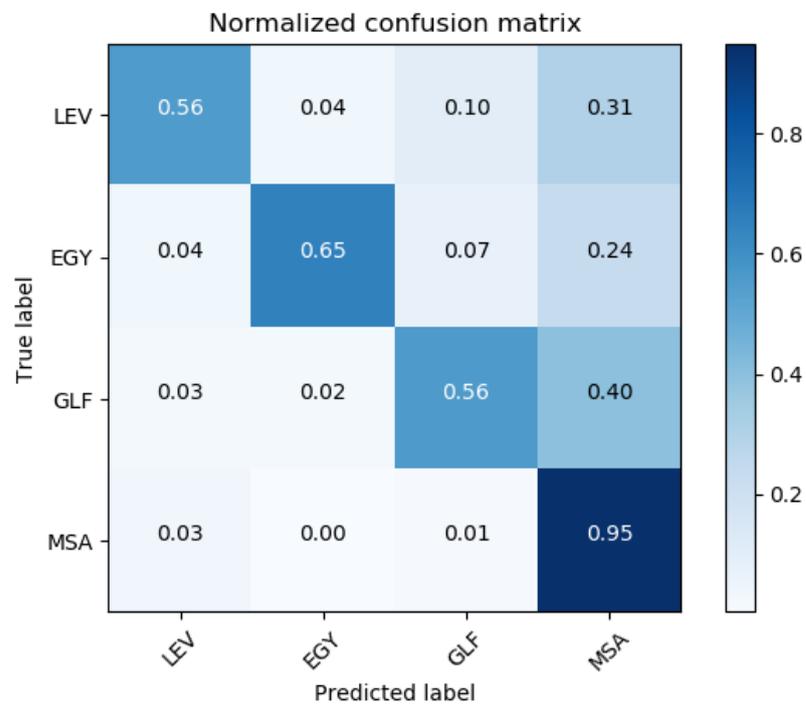

**Figure 28: BUNOC Bi-gram Model Confusion Matrix in Testing Arabic Dialect**

The confusion matrix of Arabic dialects identification on testing dataset shows that 40% of the GLF errors are confused with MSA, followed by LEV errors (confused with MSA 31% of the time) and finally EGY errors (confused with MSA 24% of the time). These results are because of the high lexical overlap between dialects and MSA, which make the Arabic language processing challenging tasks.

Table 11, Table 12 and Table 13 compares the performance of the proposed model with the introduced models results in [78] which was achieved by applied 7 experiment using traditional classifier, and 18 experiment models using Deep Learning with Random Embeddings , AOC Embeddings, and Twitter-City Embeddings using AOC dataset.

Table 11: Comparison with Group-A Model for Arabic Dialects Identification

| Outperform The Following Classifier In Val & Test | | |
|---|---|---|
| **Traditional Model** | **Val** | **Test** |
| Baseline (majority class in Train) | 46.49 | 46.49 |
| Logistic Regression (1+2+3 grams) | 75.75 | 78.24 |
| Naive Bayes (1+2+3 grams) | **80.15** | 77.75 |
| SVM (1+2+3 grams) | 74.5 | 75.82 |
| Logistic Regression (1+2+3 grams TF-IDF) | 75.81 | 78.24 |
| Naive Bayes (1+2+3 grams TF-IDF) | 73.21 | 75.81 |
| SVM (1+2+3 grams TF-IDF | 76.65 | **78.61** |
| **Deep Learning Model** | | |
| CNN – Random Embedding | 66.34 | 68.86 |
| CLSTM – Random Embedding | 64.58 | 65.25 |
| LSTM – Random Embedding | 70.21 | 68.71 |
| BiLSTM – Random Embedding | 75.94 | 77.55 |
| BiGRU – Random Embedding | 74.56 | 76.51 |
| CNN – AOC Embedding | 64.23 | 64.17 |
| CLSTM – AOC Embedding | 64.61 | 63.89 |
| LSTM – AOC Embedding | 70.01 | 68.91 |
| BiLSTM – AOC Embedding | 76.12 | 78.35 |
| CNN – Twitter Embedding | 74.13 | 75.61 |
| CLSTM – Twitter Embedding | 79.41 | 77.51 |
| LSTM – Twitter Embedding | 75.21 | 78.53 |
| **Our Model** | | |
| BUNOC Bi-gram @ 39 Epoch | **81.59** | **79.76** |
| BUNOC Tri-gram @ 52 Epoch | **81.07** | **79.47** |
| BUNOW @ 33 Epoch | **80.52** | 78.41 |
| BUNOW @ 10 Epoch | 78.62 | 77.00 |

Table 12: Comparison with Group-B Model for Arabic Dialects Identification

| Outperform The Following Classifier In Val Only | | |
|---|---|---|
| **Deep Learning Model** | **Val** | **Test** |
| Attention-BiLSTM – Random Embedding | 79.97 | 80.21 |
| BiGRU – AOC Embedding | 79.61 | 80.11 |
| Attention-BiLSTM – AOC Embedding | 80.25 | **81.12** |
| **Our Model** | | |
| BUNOC Bi-gram @ 39 Epoch | **81.59** | 79.76 |
| BUNOC Tri-gram @ 52 Epoch | **81.07** | 79.47 |
| BUNOW @ 33 Epoch | **80.52** | 78.41 |
| BUNOW @ 10 Epoch | 78.62 | 77.00 |

**Table 13: Comparison with Group-C Model for Arabic Dialects Identification**

| Classifier Cannot Outperform In Both Val & Test | | |
|---|---|---|
| **Deep Learning Model** | **Val** | **Test** |
| BiLSTM – Twitter Embedding | 82.81 | 81.93 |
| BiGRU – Twitter Embedding | 83.25 | 82.21 |
| Attention-BiLSTM – Twitter Embedding | **83.49** | **82.45** |
| **Our Model** | | |
| BUNOC Bi-gram @ 39 Epoch | 81.59 | 79.76 |
| BUNOC Tri-gram @ 52 Epoch | 81.07 | 79.47 |
| BUNOW @ 33 Epoch | 80.52 | 78.41 |
| BUNOW @ 10 Epoch | 78.62 | 77.00 |

# Chapter 6: Discussion

Text data is growing rapidly because of internet evolution. Its became mandatory to develop a powerful tools able to automatically handle this widely available of text data generated from public reviews, opinions, comments, recommendations, ratings, feedback, attitudes, emotions, and feelings in order to transform it into organized knowledge.

Deep learning i.e. CNN is one of best solution for automatically learn from scratch using the raw text data directly with less or no preprocessing and don't need any interfere from human for feature selection or extraction compared to traditional ML

Currently, CNN model facing the following drawback:

1. Depend on word or character of the language to encoding text data. (Language Dependent)

2. Large input feature vector especially when using a word encoding.

3. Initialize the model require too much network weight parameter #

4. Difficult in applying to CJK language (Chinese, Japanese, Korean) without Romanization to English, as those language native character letter > 4,000

5. Lack of research on Arabic textual dialect identification to solve the problem of misclassification due to high lexical overlap between the dialects and MSA

The works under this research aim to enhancement the automatically text classification using CNN and overcome the drawback of the existence Techniques by introducing a proposed model that is characterize with the following:-

1. Be language independent.

2. Low level representation of text data which directly affect the memory space consumption by reducing the input feature vector and its consequent CNN network weight parameters #

3. Efficiency in handling the textual data, where the predefined classes are interrelated with each other i.e. Arabic dialect identification, in which the dialects and MSA are lexical overlap.

This thesis introduced an innovative text encoding method, which is able to work on word level and character N-gram level for encoding raw text data by converting each N-gram unique dictionary integer ID number to its equivalent binary value. We also proposed a new spatial 2D CNN architecture with horizontal convolutional and vertical convolutional to be compatible and efficiency with our encoding method.

For measure, our proposed model achievement against defined target objectives. We used two benchmarked datasets from two different morphological languages, as one belongs to Latin script (English language) for text classification while other belongs to Arabic script (Arabic language) for dialects identification. In English language, our models achieved a competitive accuracy compared to state of art models in addition to reducing input feature vector, and decreasing memory consumption space. In Arabic dialects identification, our models achieved competitive accuracy results also using only the in-domain training dataset and randomly initialed neural network weights compared with other models using different deep learning architecture in addition to three different embedding settings.

Our results show the powerful of using convolutional neural network and its ability to outperform the most complicated deep neural network models with in-domain training dataset without the need of pre-trained word2vec or embedding layer. One of the major observed from our experiment that is the way of feeding the raw data into CNN has highly effect on the performance and accuracy of the model.

# Chapter 7: Conclusions & Future Work

This thesis presents a two new encoding methods "BUNOW" and "BUNOC" used for feeding the raw text data into spatial CNN architecture instead of commonly used methods like one hot vector or word representation (i.e. word2vec) with temporal CNN architecture. The main core idea depends on representing each N-gram as a fixed k-dimensional binary vector equivalent to the unique integer ID of this word inside the dictionary.

Our model is language-independent with small input feature vector, which allows less number of neural network parameters. The proposed method can be classified as hybrid word-character models in its work methodology because it consumes less memory space by using a fewer neural network parameters as in character level representation, in addition to providing much faster computations with fewer network layers depth, where a word is an atomic representation of the document as in word level.

Our results have been evaluated in two different morphological languages by using two-benchmarked dataset one for Arabic language and one for English language.

In English AG's News dataset, the provided CNN model outperforms the character level and very deep character level CNNs in terms of accuracy, network parameters, and memory consumption. Compared to 26 research experiments used the same dataset, the proposed model achieved 8.01% test error, which is lower than any of the recorded results achieved by the current CNN models. Additionally, it is reducing the input feature vector and neural network parameters by 62% and 34%, respectively.

In Arabic AOC dataset, the proposed model with randomly initialized neural network weights achieved competitive accuracy results of 81.59% in validation and 79.76% in testing compared to the best result of 80.15% and 78.61 achieved from 7 traditional machine-learning classifiers, and 12 different deep learning architectures. It depends only on the in-domain dataset compared with three different techniques to initialized deep neural network used by other researchers where out-domain pre-trained word2vec was introduced.

Despite the promising results achieved by the proposed model, many directions will be explored in the future work:

**Efficiency Measurement:** Due to the limitation of resources and time, this thesis not study neither the efficiency on large-scale dataset nor training and testing time of proposed model against the state of art models.

**Feature Engineering:** Study the effect of intensively preprocessing, and using feature selection and extraction methods on accuracy results.

**Different Deep Learning Architectures:** Although the experiments make use of CNN to produce competitive results. Other architectures could also be used in future work, such as LSTM, CLSTM, GRU, and attention mechanism.

**Multilingual Text:** Although the proposed model is, characterize as language independent and evaluated on two different morphological language dataset in which a single language approach used for each language separately from the other. Its performance on multilingual text classification has not been tested.